\documentclass[lettersize,10pt,journal]{IEEEtran}
\usepackage{amsmath,amsfonts}
\usepackage{array}
\usepackage{textcomp}
\usepackage{stfloats}
\usepackage{url}
\usepackage{verbatim}
\usepackage{cite}
\usepackage{times}
\usepackage{xcolor}

\usepackage{graphics} 
\usepackage{subcaption}
\usepackage{epsfig} 
\usepackage{times} 
\usepackage{amsmath} 
\usepackage{mathtools,amsfonts}
\usepackage{amssymb}  
\usepackage[T1]{fontenc}
\DeclareTextSymbolDefault{\dh}{T1}
\usepackage{hyperref}
\usepackage{url}

\usepackage{algorithm}
\usepackage[noend]{algpseudocode}
\usepackage{newtxtext}
\usepackage{newtxmath}
\usepackage{lettrine}
\usepackage{booktabs}
\usepackage{multirow}
\usepackage{graphicx}
\usepackage{multicol}
\usepackage{caption}
\usepackage{mwe}
\usepackage{hyperref}
\usepackage{placeins}
\usepackage{svg}
\usepackage{booktabs}
\usepackage{array}
    \newcolumntype{P}[1]{>{\centering\arraybackslash}p{#1}}
    \newcolumntype{M}[1]{>{\centering\arraybackslash}m{#1}}

\hypersetup{
    colorlinks=true,
    linkcolor=blue,
    filecolor=magenta,      
    urlcolor=cyan,
}

\DeclarePairedDelimiter\abs{\lvert}{\rvert}%
\DeclarePairedDelimiter\norm{\lVert}{\rVert}%

\makeatletter
\let\oldabs\abs
\def\abs{\@ifstar{\oldabs}{\oldabs*}}
\let\oldnorm\norm
\def\norm{\@ifstar{\oldnorm}{\oldnorm*}}
\makeatother

\definecolor{deep-red}{RGB}{192, 0, 0}
\definecolor{deep-purple}{RGB}{120, 0, 170}
\definecolor{good-green}{RGB}{0,175,0}
\definecolor{purple}{RGB}{210, 0, 210}
\definecolor{alizarin}{rgb}{0.82, 0.1, 0.26}

\usepackage[textsize=scriptsize, bordercolor=white,backgroundcolor=gray!30,linecolor=black,colorinlistoftodos]{todonotes}  
\usepackage[normalem]{ulem} 
\usepackage{soul} 

\usepackage[mathscr]{euscript}
\newif\ifTrackChanges   
\TrackChangesfalse 
\ifTrackChanges
    
    \newcommand{\NStrike}[1]{\textcolor{alizarin}{\st{#1}}}
    
    \newcommand{\HStrike}[1]{\textcolor{purple}{\st{#1}}}
    \newcommand{\cut}[1]{{\color{gray}{#1}}}
\else
    
    \newcommand{\NStrike}[1]{}
    \newcommand{\HStrike}[1]{}
    
    \newcommand{\cut}[1]{{}}
\fi

\newcommand*\circled[1]
{\tikz[baseline=(char.base)]{\node[circle,minimum size=6pt,draw=black,inner sep=0.25pt](char){\scriptsize #1};}}
\begin{document}

\title{Long Horizon Planning through Contact using Discrete Search and Continuous Optimization}

\author{Ramkumar Natarajan$^{1}$, Garrison L.H. Johnston$^{2}$, Nabil Simaan$^{2}$, Maxim Likhachev$^{1}$ and Howie Choset$^{1}$
\thanks{$^{1}$ The authors are with The Robotics Institute at Carnegie Mellon University, Pittsburgh PA 15213. email: \{\tt\small rnataraj, maxim, choset\}@cs.cmu.edu }
\thanks{$^{2}$ The authors are with the department of Mechanical Engineering at Vanderbilt University. email: \{\tt\small garrison.l.johnston, nabil.simaan\}@vanderbilt.edu }
\thanks{This work was supported by NSF awards \#1734461 and \#1734460, ARL grant W911NF-18-2-0218 and by Vanderbilt and Carnegie Mellon internal university funds.}
}

\markboth{}{Natarajan \MakeLowercase{\textit{et al.}}: Torque-Limited Manipulation Planning through Contact using INSAT}

\renewcommand\intercal{{\cramped{{}^\mathsf{T}}}}

\maketitle

\begin{abstract}

Robots often have to perform manipulation tasks in close proximity to people (Fig. \ref{fig:motiv}). As such, it is desirable to use a robot arm that has limited joint torques to not injure the nearby person and interacts with the environment to explore new possibilities for completing a task. By bracing against the environment, robots can expand their reachable workspace, which would otherwise be inaccessible due to exceeding actuator torque limits, and accomplish tasks beyond their design specifications. However, motion planning for complex contact-rich tasks requires reasoning through the permutations of different possible contact modes and bracing locations, which grow exponentially with the number of contact points and links in the robot. To address this combinatorial problem, we developed INSAT \cite{insat_ptc}, which interleaves graph search to explore the manipulator joint configuration and the contact mode space with incremental trajectory optimizations seeded by neighborhood solutions to find a dynamically feasible trajectory through contact. In this paper, we present recent additions to the INSAT algorithm that improve its runtime performance. In particular, we propose Lazy INSAT with reduced optimization rejection that systematically procrastinates its calls to trajectory optimization while reusing feasible solutions that violate boundary constraints. The algorithm is evaluated on a heavy payload transportation task in simulation and on physical hardware. In simulation, we show that Lazy INSAT can discover solutions for tasks that cannot be accomplished within its design limits and without interacting with the environment. In comparison to executing the same trajectory without environment support, we demonstrate that the utilization of bracing contacts reduces the overall torque required to execute the trajectory.

\end{abstract}

\section{Introduction}


\lettrine{C}{}ollaborative robots play a crucial role in alleviating the physical strain associated with demanding tasks for human operators working in confined spaces. These robots assist by handling heavy payloads within the confines of restricted areas. For such tasks, these robots require massive long-reach links and large torque actuators to support their own weight and the payload. However, fulfilling such operational requirements poses challenges to the safety of human-robot collaboration in close proximity. Consequently, we encounter a manipulation planning problem wherein the planner must minimize manipulator joint torques and accelerations, uphold task manipulation requirements, and navigate around obstacles.

\begin{figure}[h!]
\centering
\includegraphics[width=\columnwidth]{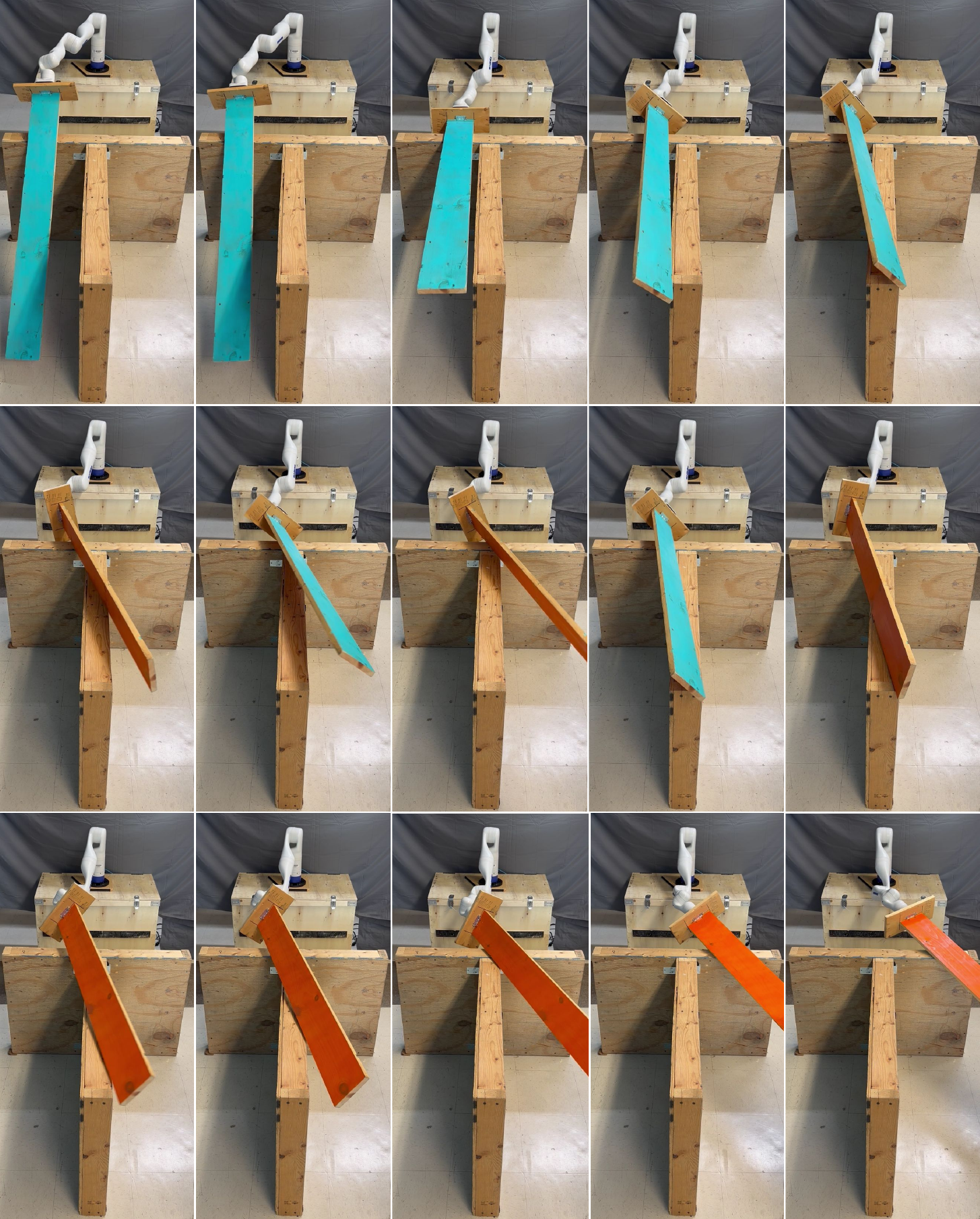}
\caption{Kinova Gen3 manipulator transferring a long heavy wooden plank across a partitioned space by bracing against the environment for support.}\label{fig:realflip_intro}
\end{figure}


\par To address the conflicting demands of safe collaboration and operation in confined spaces, it has been demonstrated that robots can strategically brace against the environment, thereby reducing the overall effort required to manipulate heavy objects \cite{braceadv}. The physical constraints imposed by the environment can be transformed into opportunities, enabling efficient manipulation that consumes less energy, enhances accuracy \cite{Hollis1992Bracing,bracing2}, and reduces compliance \cite{Johnston2020Bracing}. 
\par Recently, we introduced a motion planning algorithm for manipulation that autonomously identifies and leverages bracing locations along the entire trajectory to achieve a desired contact-rich task \cite{insat_ptc}. Consequently, our torque-limited manipulation planning algorithm opportunistically makes/breaks/sustains contact with the environment, allowing the robot to reach deep within confined spaces with insufficient actuator torques or carry heavy payloads beyond the manipulator's inherent capacity. The algorithm presented in our previous work can be summarized as follows:
\begin{itemize}
    \item A novel adaptation of INSAT: INterleaved Search And Trajectory optimization \cite{insat} for the application of torque-limited manipulation planning through contact. By \textit{interleaving} discrete graph-search with continuous trajectory optimization, our algorithm planned through contact over long horizons for high-dimensional complex manipulation problems in confined non-convex environments.
    \item A dynamically feasible trajectory through contact can be non-smooth with impacts and discontinuities. We proposed a new virtual contact frictional force model to enable planning for complex, contact-rich motions without relying on a pre-specified contact schedule using a gradient-based optimizer.
    \item To the best of our knowledge, manipulation planning that actively reasoned about effort reduction by utilizing additional support from contact has not been proposed and demonstrated on a real robot arm before \cite{insat_ptc}. Hence this formed the most important contribution of \cite{insat_ptc}.
\end{itemize}

\textbf{Statement of Extension:} The updated algorithm presented here evolved from the conference version \cite{insat_ptc} in the four following ways:
\begin{itemize}
    \item Based on the observation of the complexity of different re-wiring operations in INSAT, we propose a lazy version of INSAT that delays evaluating long-horizon trajectory optimizations without sacrificing any of the properties of INSAT. 
    \item We extend INSAT to utilize a discrete seed path that is not dynamically feasible. This seed path is also generated within the algorithm and dramatically increases the runtime speed of INSAT. 
    \item Using risk-sensitive cost function transformation and other numerical techniques, we are able to simplify the trajectory optimization when compared to \cite{insat_ptc}.  
    \item One of the biggest limitations of trajectory optimization methods is their inability to work reliably for long-horizon problems. However, in many cases, their solutions are feasible but do not satisfy the given boundary conditions. This work proposes a way to reuse such \textit{partial} solutions by introducing additional nodes to the low-D graph.
\end{itemize}

\begin{figure}
\centering
\includegraphics[width=\columnwidth]{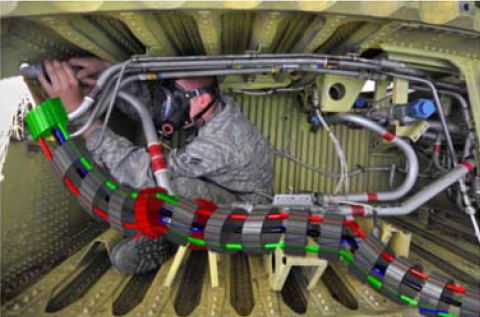}
\caption{An example of a hyperredundant robot manipulator lifting a heavy tool in a confined space by leveraging contact with the environment to assist a human worker.}\label{fig:motiv}
\end{figure}

This paper is structured as follows: we discuss prior work in Sec. \ref{sec:relwork}, formalize the problem statement in \ref{sec:ps} and introduce the tunable virtual contact models for contact-implicit trajectory optimization in \ref{sec:prelim}. We then describe our proposed method in Sec. \ref{sec:topus}. Finally, we show the experimental results in Sec. \ref{sec:results}, and conclude in Sec. \ref{sec:conc}. 

\color{black}


\section{Prior Work}
\label{sec:relwork}

Although the concept of bracing against the environment for manipulation was first proposed in the 1980s \cite{bracing1}, it has not received significant attention since then. Sensing and control for bracing, incorporating probabilistic contact estimation \cite{estbrace} and handling multiple simultaneous contacts with the environment \cite{contbrace}, were introduced in the 2000s. The methods proposed in \cite{planbrace} and \cite{braceadv} in the late 2010s were the first to consider manipulation planning through bracing against the environment. However, \cite{braceadv} primarily serves as a control algorithm to guide the robot to the right contact point and posture, rather than a planner that reasons about bracing to achieve a desired long-term task. In contrast, the idea presented in \cite{planbrace} is limited as it (i) ignores contact dynamics, (ii) only considers static environmental bracing, and (iii) is demonstrated only on a simple planar elastomer manipulator with a single 2D obstacle.

In contrast, our planner generates trajectories with dynamic bracing (contact sliding), is evaluated in a much more challenging environment, and is demonstrated on a real robot. Planning to brace and navigating through the environment by bracing can be interpreted as constrained manipulation planning over a sub-manifold, a concept addressed by \cite{cbirrt} and its variants. However, these quasi-static methods do not account for the dynamics of contact and the manipulator system that are crucial for understanding and leveraging the effects of bracing.

Planning through contact with an unspecified mode sequence is an active challenge in robot locomotion and manipulation \cite{cito}. The goal of the general formulation, called Contact Implicit Trajectory Optimization (CITO), is to jointly find trajectories for state, control input, and contact forces. Previous successful works propose different combinations of trajectory optimization-based approaches \cite{citoonol}, including direct shooting \cite{citoshoot} and direct transcription \cite{timestep, timestepmulti}. To incorporate contact, they use either complementarity conditions with implicit time-stepping \cite{timestep, timestepmulti} or soft constraints implemented as a penalty term \cite{penalty1, penalty2} in the cost function \cite{cio1, cio2}. However, standalone optimization-based approaches are fragile when it comes to global reasoning over a long horizon and depend heavily on the quality of initial guesses.

On the other hand, the contact mode sequence is inherently discrete, and the optimizer faces a fundamentally discrete choice at each time, which is challenging to optimize, whether modeled using continuous constraints or integer variables. Recent approaches use graph search-based methods \cite{contsearch1, contsearch2, contsearch3} or rapidly-exploring random trees \cite{contrrt} to plan contact switches and generate a seed for the subsequent trajectory optimization. However, these local methods are greedy and do not offer a fallback in case the trajectory optimization does not succeed using the discrete contact sequence. In contrast, INSAT provides a principled way to globally reason over the discrete and continuous parts of the problem.


\section{Problem Statement}
\label{sec:ps}

In this work, we designate the robot manipulator as $\mathcal{R}$, with $\mathcal{X}^\mathcal{R} \subseteq \mathbb{R}^N$ representing the configuration space (C-space) for an \(N\)-degree-of-freedom (DoF) manipulator. Let $\mathcal{X}^{\text{obs}} \subset \mathcal{X}^\mathcal{R}$ denote the C-space obstacle, $\mathcal{X}^{\text{free}} = \mathcal{X}^\mathcal{R} \setminus \mathcal{X}^{\text{obs}}$ as the free space, and $\mathcal{X}^S \subset \mathcal{X}^{\text{obs}}$ representing the surface of the obstacle with which the robot can make and break contact. The planning state is composed of joint angles and joint velocities, denoted as $\textbf{x} = [\textbf{q}, \textbf{\.q}] \in \mathcal{X} \subseteq \mathbb{R}^{2N}$. The manipulator is controlled by bounded joint torque inputs, given by $\textbf{u} \in \mathbb{R}^N$. Given (a) a start state $\textbf{x}^S$, (b) a goal state $\textbf{x}^G$, (c) the planning space $\mathcal{X}$ with obstacles $\mathcal{X}^{\text{obs}}$ and the obstacle contact surface $\mathcal{X}^S$, the task is to find a control trajectory $\textbf{u}(t)$ for \(t \in [0,T]\) according to Eq. \ref{eq:obj}.

In many manipulation planning problems, the state space $\mathcal{X}$ can be high-dimensional. Let us consider a low-dimensional subspace $\mathcal{X}_L$ of the state space $\mathcal{X}$, such that $\mathcal{X}_L \subset \mathcal{X}$. An invertible many-to-one mapping $\boldsymbol{\lambda}: \mathcal{X} \longrightarrow \mathcal{X}_L$ is introduced to project a full-dimensional state $\textbf{x} = [\textbf{q}, \dot{\textbf{q}}] \in \mathcal{X}$ into the low-dimensional space $\mathcal{X}_L$. This mapping is defined as $\textbf{x}_L = \boldsymbol{\lambda}(\textbf{x})$. The inverse mapping, denoted by $\boldsymbol{\lambda}^{-1}: \mathcal{X}_L \longrightarrow \mathcal{X}$, is a one-to-many mapping that lifts a low-dimensional state $\textbf{x}_L \in \mathcal{X}_L$ to any possible full-dimensional state $\textbf{x} \in \mathcal{X}$. Therefore, $\textbf{x} = \boldsymbol{\lambda}^{-1}(\textbf{x}_L)$. The term $\boldsymbol{\phi}_{\textbf{x}^\prime\textbf{x}^{\prime\prime}}(t)$ denotes a time \(t\)-parameterized full-dimensional trajectory from $\boldsymbol{\lambda}^{-1}(\textbf{x}^\prime_L)$ to $\boldsymbol{\lambda}^{-1}(\textbf{x}^{\prime\prime}_L)$. The argument \(t\) may be dropped for brevity.
\par For torque-limited planning, the manipulator's maximum velocity and torque constraints must be satisfied while planning. The energy-optimal motion planning for torque-limited manipulation can be cast as the following optimization:
\begin{equation}
\begin{aligned}
\text{find} \ \ & \textbf{u}(t) \\
\text { s.t. } & \boldsymbol{\phi}_{\textbf{x}^S\textbf{x}^G}(t)=\textbf{f}(\textbf{x}^S, \textbf{u}(t)), \\
& \boldsymbol{\phi}_{\textbf{x}^S\textbf{x}^G}(T) = \textbf{x}^G \\
& \boldsymbol{\phi}_{\textbf{x}^S\textbf{x}^G}(t) \in (\mathcal{X} \setminus \mathcal{X}^{\text{obs}}) \cup \mathcal{X}^S\\
& |\boldsymbol{\dot{\phi}}_{\textbf{x}^S\textbf{x}^G}(t)| \leq \dot{\textbf{x}}_{\text{lim}}, |\boldsymbol{\ddot{\phi}}_{\textbf{x}^S\textbf{x}^G}(t)| \leq \ddot{\textbf{x}}_{\text{lim}}, |\textbf{u}(t)| \leq \textbf{u}_{\text{lim}} \\
\end{aligned}
\label{eq:obj}
\end{equation}
where $\textbf{f}$ denotes the manipulator dynamics with contact that captures the interaction of $\mathcal{R}$ with the environment (Eq. \ref{eq:manipdyn}), $\dot{\textbf{x}}_{\text{lim}}$ and $\ddot{\textbf{x}}_{\text{lim}}$ are limits on velocity and acceleration respectively. $\boldsymbol{\phi}_{\textbf{x}^S\textbf{x}^G}(t)$ is the trajectory connecting start $\textbf{x}_S$ to goal $\textbf{x}_G$ computed by forward simulating the dynamics starting from $\textbf{x}_S$ using $\textbf{u}(t)$. Note that $\boldsymbol{\phi}_{\textbf{x}^S\textbf{x}^G}(t)$ can lie on $\mathcal{X}^S$ and hence encodes the sequence of making and breaking contact with the environment.


\section{Background on INSAT}
\subsection{A* search}
\label{sec:astar}
Before introducing INSAT, the reader can benefit from a quick summary of the A* search. A* or its variant is the underlying search algorithm INSAT uses to explore the state space and generate dynamically feasible trajectories. The simplest version of A* search is shown in Alg. \ref{alginsat:astar}. A* builds a graph that approximates the feasible space and searches over it to find a path from start $\textbf{x}_L^S$ to goal $\textbf{x}_L^G$ (Alg. \ref{alginsat:astar}: line \ref{lineinsat:a:succ}). At every iteration, a node is selected from the frontier of the graph for expansion, generating its successors and forming a new frontier. The node picked for expansion is based on a priority cost (Alg. \ref{alginsat:astar}: line \ref{lineinsat:a:pq}) which is a sum of two costs: $c_\text{come}$ and $c_\text{go}$. $c_\text{come}(\textbf{x}_L^\prime)$ tracks the best cost yet to reach $\textbf{x}_L^\prime$ from start and $c_\text{go}(\textbf{x}_L^\prime)$ is an estimate of cost-to-go to goal from $c_\text{come}(\textbf{x}_L^\prime)$. Upon expanding a node, if the successor is visited via a better parent, then the new parent and cost-to-come value of the successor are updated (Alg. \ref{alginsat:astar}: lines \ref{line:a:update1}-\ref{line:a:update2}). A* guarantees provably optimal solution cost by selecting $c_\text{go}(\textbf{x}_L^\prime)$ to be an underestimate of the optimal cost-to-go value. 
    
\setlength{\textfloatsep}{4pt}
\begin{algorithm}
\begin{algorithmic}[1]
\Procedure{Main}{$\textbf{x}^{S}_L, \textbf{x}^{G}_L$}
\State $\textbf{x}^{\text{next}}_L = \textbf{x}^S_L$ 
\State \textbf{while} $\textbf{x}^G_L$ is not expanded \textbf{do} \label{lineinsat:a:termin}
\State $\>$ $\textbf{x}^{\text{next}}_L$ = $\arg\min_{\textbf{x}_L \text{in frontier}} \{c_{\text{come}}(\textbf{x}_L) + c_{\text{go}}(\textbf{x}_L)\}$ \label{lineinsat:a:pq}
\State $\>$ Generate all the valid successors $\textbf{X}^{\text{new}}_L$ of $\textbf{x}^{\text{next}}_L$  \label{lineinsat:a:succ}
\State $\>$ \textbf{for} $\textbf{x}^{\text{new}}_L$ in $\textbf{X}^{\text{new}}_L$ \textbf{do}
\State $\>$ $\>$ \textbf{if} $c_{\text{come}}(\textbf{x}^{\text{new}}_L) > c_{\text{come}}(\textbf{x}^{\text{next}}_L) + c(\textbf{x}^{\text{next}}_L, \textbf{x}^{\text{new}}_L)$ \textbf{then} \label{line:a:update1}
\State $\>$ $\>$ $\>$ $c_{\text{come}}(\textbf{x}^{\text{new}}_L) = c_{\text{come}}(\textbf{x}^{\text{next}}_L) + c(\textbf{x}^{\text{next}}_L, \textbf{x}^{\text{new}}_L)$
\State $\>$ $\>$ $\>$ Set $\textbf{x}^{\text{next}}_L$ as parent of $\textbf{x}^{\text{new}}_L$ \label{line:a:update2}
\EndProcedure
\end{algorithmic}
\caption{A* Search}
\label{alginsat:astar}
\end{algorithm}

\begin{figure*}[h!]
    \centering
    \includegraphics[width=\textwidth]{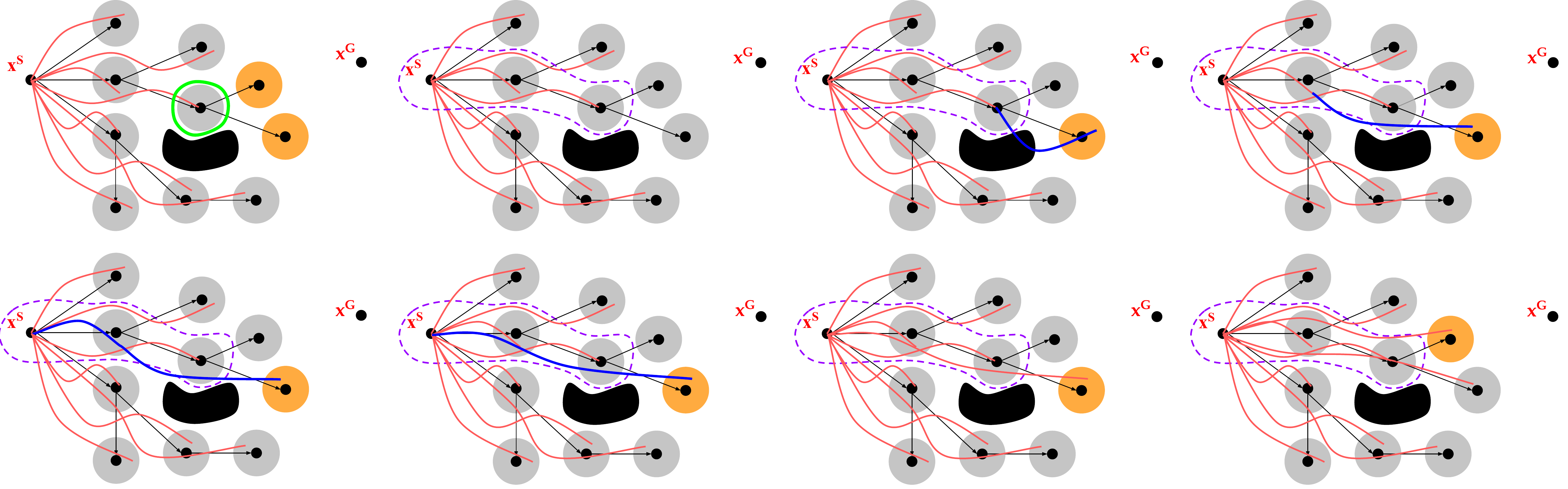}
    \caption{One iteration of node expansion, successor generation and full-D trajectory optimization in INSAT. Here $\textbf{x}^S$ and $\textbf{x}^G$ are full-D start and goal states. The low-D graph is denoted using black dots and arrows as nodes and edges. The one-to-many mapping of low-D node to full-D space is shown using the gray circle around the low-D node. The red curves denote the full-D dynamically feasible trajectory. ROW MAJOR ORDER: (1) An intermediate step in INSAT algorithm (2) INSAT selects the node to be expanded (encircled in green) (3) and generates the low-D successors (orange nodes). (4) INSAT searches over all the ancestors of the newly generated successor (encircled with a dotted purple line) to find a valid full-D trajectory to the successor. (5)-(6) Invalid trajectory from some of the ancestors. (7)-(11) Once a valid trajectory is found, the trajectory optimization is warm-started with the incoming trajectory of the valid ancestor and the incremental trajectory found in the previous step to find the valid full-D dynamically feasible trajectory from start $\textbf{x}^S$. (12) The same process is repeated for the other newly generated successor.} 
    \label{figinsat:insatgeneric}
\end{figure*}

\subsection{Key Idea of INSAT}
\begin{figure}[h!]
    \centering
    \includegraphics[width=\columnwidth]{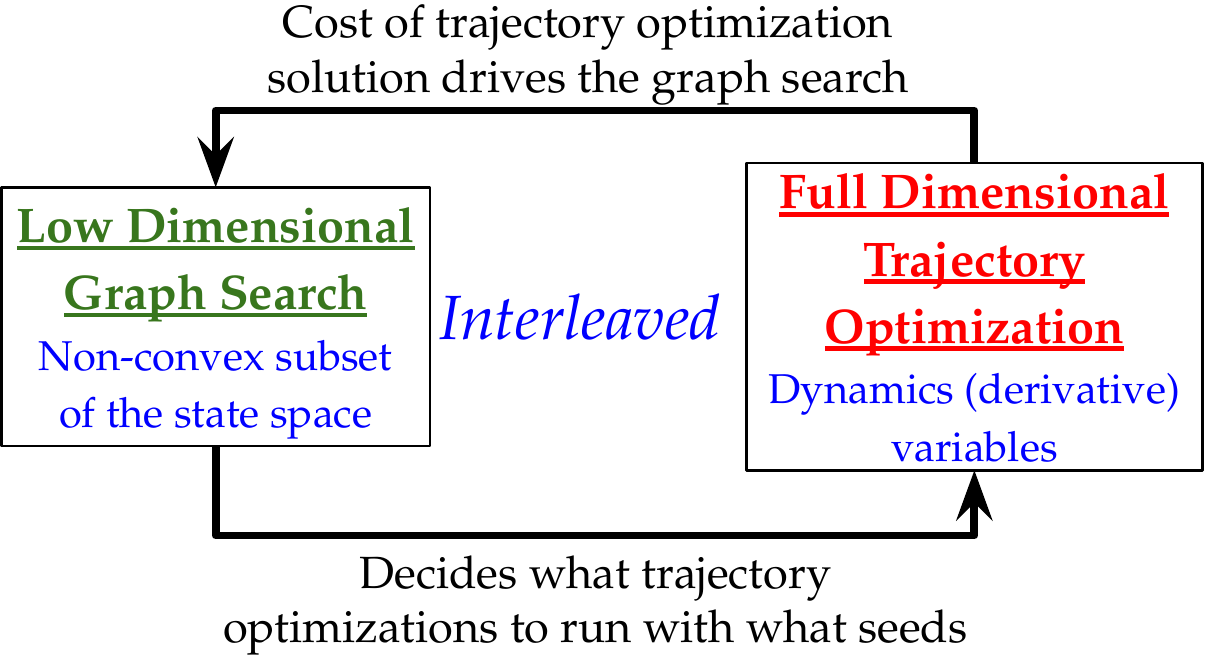}
    \caption{Working principle of INSAT}
    \label{figinsat:schematic}
\end{figure}

The key idea behind INSAT is (a) to identify a low-dimensional manifold, (b) perform a search over a grid-based or sampling-based graph that represents this manifold, (c) while searching the graph, utilize high-dimensional trajectory optimization to compute the cost of partial solutions found by the search (see Fig. \ref{figinsat:schematic}). As a result, the search over the lower-dimensional graph decides what trajectory optimizations to run and with what seeds, while the cost of solution from the trajectory optimization drives the search in the lower-dimensional graph until a feasible high-dimensional trajectory from start to goal is found. The dynamic programming in the low-dimensional search enables warm-starting trajectory optimizations using highly informative initial guesses.
\subsection{Low Dimensional Graph Search}
As such, INSAT is agnostic to the choice of the low-dimensional graph search algorithm. For the low-dimensional search, any graph search-based algorithms like A* \cite{astar}, wA* \cite{pohlwastar}, ARA* \cite{maxara}, MHA* \cite{mha} or sampling-based algorithms like RRT \cite{rrt} and its variants can be used. To guarantee completeness in INSAT using discrete graph search methods in low-D, the discretization should be chosen such that there exists at least a path from $\textbf{x}^S$ to $\textbf{x}^G$ on the low-D graph whose full-D subspace contains the full-D solution trajectory $\boldsymbol{\phi}_{\textbf{x}^S\textbf{x}^G}$.


\subsection{Full Dimensional Trajectory Optimization}
INSAT is also agnostic to the choice of the full-dimensional trajectory optimization method. Indirect shooting methods \cite{citoonol} methods are preferred as they offer dynamic feasibility in every iteration during convergence. As the choice of trajectory optimization method is generally tied to the type of system the trajectory is optimized for, it is advised that the practitioner exploit this structure. For instance, for differentially flat systems, one need not use a fully nonlinear solver to solve the original trajectory optimization problem. Instead, they can be converted to a quadratic program (QP) over piecewise polynomials by exploiting their dynamic inversion and can be solved significantly faster \cite{roypoly}. 


\setlength{\textfloatsep}{4pt}
\begin{algorithm}
\begin{algorithmic}[1]
\Procedure{Main}{$\textbf{x}^{S}, \textbf{x}^{G}$}
\State $\textbf{x}^{\text{next}}_L = \textbf{x}^S$ 
\State \textbf{while} $\boldsymbol{\phi}_{\textbf{x}^{\text{S}}\textbf{x}^{\text{G}}}$ is EMPTY \textbf{do} \label{lineinsat:termin}
\State $\>$ Pick the next node $\textbf{x}^{\text{next}}_L$ to expand \Comment{Sec. \ref{sec:astar}} \label{lineinsat:pq}
\State $\>$ Generate the successors $\textbf{X}^{\text{new}}_L$ of $\textbf{x}^{\text{next}}_L$ \Comment{Sec. \ref{sec:astar}} \label{lineinsat:succ}
\State $\>$ \textbf{for} $\textbf{x}^{\text{new}}_L$ in $\textbf{X}^{\text{new}}_L$ \textbf{do}
\State $\>$ $\>$ Get the ancestors $\textbf{X}^{\text{pred}}_L$ of $\textbf{x}^{\text{new}}_L$ \label{lineinsat:ances}
\State $\>$ $\>$ \textbf{for} $\textbf{x}^{\text{pred}}_L$ in $\textbf{X}^{\text{pred}}_L$ \textbf{do}
\State $\>$ $\>$ $\>$ $\boldsymbol{\phi_{\textbf{x}^{\text{pred}}\textbf{x}^{\text{new}}}}$ = \textproc{TrajOpt}($\textbf{x}^{\text{pred}}_L$, $\textbf{x}^{\text{new}}_L$) \label{lineinsat:trajopt}
\State $\>$ $\>$ $\>$ \textbf{if} $\boldsymbol{\phi_{\textbf{x}^{\text{pred}}\textbf{x}^{\text{new}}}}$ is VALID \textbf{then}
\State $\>$ $\>$ $\>$ $\>$  $\boldsymbol{\phi_{\textbf{x}^{S}\textbf{x}^{\text{new}}}}$ = \textproc{WarmStartOpt} \label{lineinsat:warmopt}
($\boldsymbol{\phi_{\textbf{x}^{S}\textbf{x}^{\text{pred}}}}$, $\boldsymbol{\phi_{\textbf{x}^{\text{pred}}\textbf{x}^{\text{new}}}}$) 
\State $\>$ $\>$ $\>$ $\>$ \textbf{if} $\boldsymbol{\phi_{\textbf{x}^{S}\textbf{x}^{\text{new}}}}$ is VALID \textbf{and} c($\boldsymbol{\phi_{\textbf{x}^{S}\textbf{x}^{\text{new}}}}$) < c($\textbf{x}^{\text{new}}_L$) \textbf{then}
\State $\>$ $\>$ $\>$ $\>$ $\>$ c($\textbf{x}^{\text{new}}_L$) = c($\boldsymbol{\phi_{\textbf{x}^{S}\textbf{x}^{\text{new}}}}$)
\State $\>$ $\>$ $\>$ $\>$ $\>$ Set $\textbf{x}^{\text{pred}}_L$ as the parent of $\textbf{x}_L^{\text{new}}$
\State $\>$ $\>$ $\>$ $\>$ $\>$ Store $\boldsymbol{\phi_{\textbf{x}^{S}\textbf{x}^{\text{new}}}}$
\State \textbf{return} $\boldsymbol{\phi}_{\textbf{x}^{\text{S}}\textbf{x}^{\text{G}}}$
\EndProcedure
\end{algorithmic}
\caption{INSAT}
\label{alginsat:insat}
\end{algorithm}

\subsection{INSAT Algorithm}
A minimal version of INSAT algorithm is presented in Alg. \ref{alginsat:insat} with a visual aid (Fig. \ref{figinsat:insatgeneric}). The algorithm takes the full dimensional start and goal state $\textbf{x}^S$ and $\textbf{x}^G$ as inputs and solves Eq. \ref{eq:obj}. It performs a graph search similar to the one shown in Alg. \ref{alginsat:astar} in the low-D space. But for every newly generated successor, the algorithm  runs an incremental trajectory optimization routine to find the dynamically feasible trajectory and the actual cost to reach the successor. For efficiency and speed, the trajectory optimization is warm-started by the solution already found to reach one of the ancestor nodes considered as the predecessor. This stems from the dynamic programming nature of the underlying low-D graph search.  By doing this, the algorithm finds a trajectory from start to goal subjected to non-convex cost or constraint functions and over non-convex environments resulting from the presence of obstacles. The lines \ref{lineinsat:pq} and \ref{lineinsat:succ} should be replaced with corresponding operations of the low-D search algorithm. Similarly, the lines \ref{lineinsat:trajopt} and \ref{lineinsat:warmopt} correspond to the choice of the trajectory optimization algorithm. It is worth noting that shooting methods are preferred as all the iterates towards convergence are dynamically feasible.

\section{Manipulator and Contact Model Dynamics}
\label{sec:prelim}
We model the dynamics of $\mathcal{R}$ and its interaction with the environment as a rigid-multi-body system using Euler-Lagrange mechanics with generalized coordinates $\textbf{q}$ as: 
\begin{equation}
    \textbf{M}(\textbf{q})\ddot{\textbf{q}} + \textbf{C}(\textbf{q},\dot{\textbf{q}})\dot{\textbf{q}} + \textbf{G}(\textbf{q}) = \boldsymbol{\tau} + \textbf{J}_{\boldsymbol{\Omega}}(\textbf{q})\intercal \boldsymbol{\Omega}
\label{eq:manipdyn}
\end{equation}
where \textbf{M}, \textbf{C}, \textbf{G} are mass, Coriolis and gravity matrices, $\boldsymbol{\tau}$ is the generalized input, $\textbf{J}_{\boldsymbol{\Omega}}(\textbf{q})$ is the contact Jacobian that maps $\dot{\textbf{q}}$ to the Cartesian velocities at the external contact point, and $\boldsymbol{\Omega}$ is the contact forces.
%

In this work, we use MuJoCo \cite{mujoco} to simulate the manipulator dynamics with contact at high-fidelity. Contact introduces impacts and discontinuities in the system dynamics as the contact forces (\textit{i.e.} $\boldsymbol{\Omega}$ from MuJoCo) vanish completely when not in contact and explode at the instant of making contact. A dynamically feasible control trajectory for our application might be non-smooth as the robot has to make/break/sustain contact with the environment. To optimize for such a trajectory using a gradient-based solver, we introduce two tunable smooth contact models. A smooth contact model is differentiable even at the collision event of the contact and enables faster trajectory optimization convergence. These models provide virtual forces that can be exploited in trajectory optimization to overcome the vanishing/exploding gradients of contact dynamics and enable automatic discovery of contact locations and smooth breaking of static friction. The vector of generalized joint inputs $\boldsymbol{\tau}$ can be decomposed as follows:
\begin{equation}
    \boldsymbol{\tau} = \textbf{u} - \textbf{J}_{\boldsymbol{\Gamma}}(\textbf{q})^\intercal\boldsymbol{\Gamma}(\textbf{q}, \dot{\textbf{q}})
    \label{eq:nettorq}
\end{equation}
where $\textbf{u}$ is the joint torque input, $\boldsymbol{\Gamma}(\textbf{q}, \dot{\textbf{q}}) \in \mathbb{R}^{N_{\Gamma}}$ and $\textbf{J}_{\boldsymbol{\Gamma}}(\textbf{q})$ are respectively the generalized virtual contact forces from the tunable smooth contact models (Eq. \ref{eq:normcontmod}, \ref{eq:friccontmod}, \ref{eq:normtofric}) in the contact frame and the corresponding Jacobian matrix and $N_{\Gamma}$ is number of contact pairs. $\boldsymbol{\Gamma}(\textbf{q}, \dot{\textbf{q}})$ acts on the environment in addition to the forces due to the contact mechanics from MuJoCo (\textit{i.e.} $\boldsymbol{\Omega}$). 

\begin{figure}[h]
\centering
\includegraphics[width=\columnwidth]{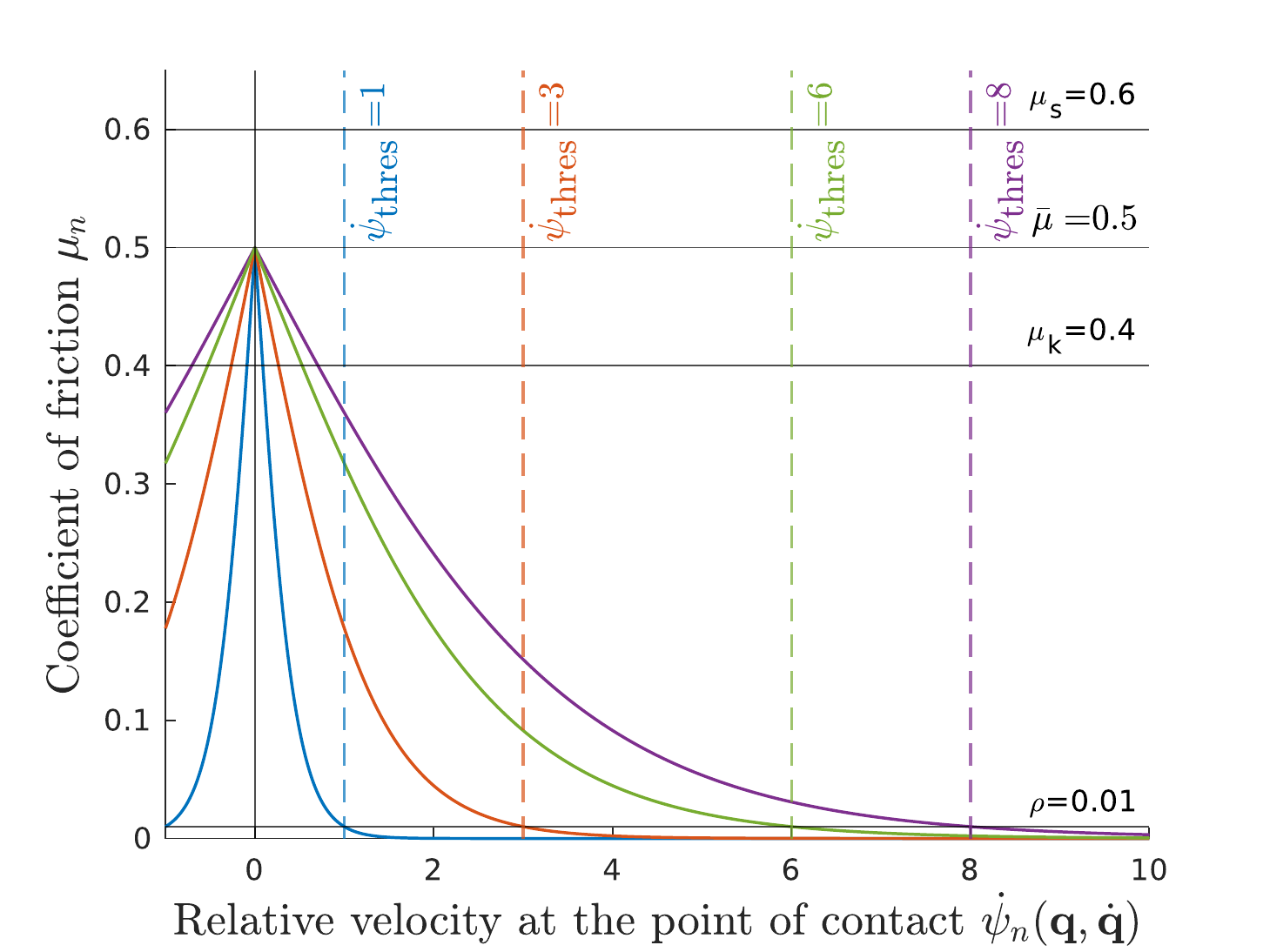}
\caption{The tunable smooth (except at $\dot{\psi}_n(\textbf{q}, \dot{\textbf{q}})=0$) contact friction model that supplies virtual frictional force to break static friction in trajectory optimization.}
\label{fig:fricmodel}
\end{figure}
\subsubsection{Tunable Smooth Contact Models} 
\label{sec:vscm}
We propose two tunable smooth contact models that supply virtual force. The first one (Eq. \ref{eq:normcontmod}) \cite{penalty2} models virtual contact normal force $\boldsymbol{\Gamma}^N_n(\textbf{q}, \dot{\textbf{q}})$ using a linear combination of nonlinear springs and dampers that resists penetration into the environment. 
\begin{equation}
\boldsymbol{\Gamma}^N_n(\textbf{q}, \dot{\textbf{q}}) = k_n e^{-\alpha_k \psi(\textbf{q})} + b_n \text{sig}(-\alpha_b \psi_n(\textbf{q})) \dot{\psi}_n(\textbf{q}, \dot{\textbf{q}})
\label{eq:normcontmod}
\end{equation}
where $\psi_n(\textbf{q})$ is the depth of penetration, $\dot{\psi}_n(\textbf{q}, \dot{\textbf{q}})$ is the relative pre-impact velocity at the point of contact, $k_n$ is the contact stiffness or spring stiffness, $b_n$ is the contact damping constant. Equation \ref{eq:friccontmod} models the virtual contact friction coefficient $\mu_n$ as a function of relative impact velocity at the point of contact (Fig. \ref{fig:fricmodel}). 
\begin{gather}
\label{eq:friccontmod}
\mu_n = \bar{\mu} - \abs{\frac{2\bar{\mu}}{1+\text{exp}\bigg(\frac{\dot{\psi}_n(\textbf{q}, \dot{\textbf{q}})}{\alpha_\mu}\bigg)} - \bar{\mu}} \\
\alpha_\mu = \frac{\dot{\psi}_{\text{thres}}}{ln\big(\frac{\rho}{2\bar{\mu}-\rho}\big)}
\end{gather}

where $\bar{\mu} = (\mu_s+\mu_k)/2$, $\mu_s$ and $\mu_k$ are the static and kinetic friction coefficients, $\mu_n$ is the virtual coefficient of friction, $\dot{\psi}_{\text{thres}}$ is the velocity threshold that breaks stiction and $\rho\to 0_+$ is a very small positive value. Then the virtual contact frictional force $\boldsymbol{\Gamma}^f_n(\textbf{q}, \dot{\textbf{q}})$ is given as
\begin{equation}
\boldsymbol{\Gamma}^f_n(\textbf{q}, \dot{\textbf{q}}) = \mu_n(\boldsymbol{\Gamma}^N_n(\textbf{q}, \dot{\textbf{q}}) + \boldsymbol{\Omega}^N)
\label{eq:normtofric}
\end{equation}

Note that in Eq. \ref{eq:normcontmod} the normal force $\boldsymbol{\Gamma}^N_n$ is nonzero and acts from a distance (\textit{i.e.} $\psi(\textbf{q}, \dot{\textbf{q}}) > 0$) when $k_n, b_n \neq 0$. By using this virtual contact force, the optimizer can discover the contact locations to brace the robot on the environment and offset the torque limits of the robot. Similarly, the coefficient of friction is equal to the average of static and kinetic friction coefficients when $\dot{\psi}_n(\textbf{q}, \dot{\textbf{q}}) = 0$. And based on the model parameter $\dot{\psi}_{\text{thres}}$ at which the object breaks static friction and starts sliding, the coefficient of virtual friction is equal to the very small value $\rho$ and the frictional force from the physics engine takes over. This enables the opposing virtual friction to counteract the static friction from the physics engine and automatically discover sliding between the objects.

The trajectory optimization is set up with costs on the tunable parameters of the smooth virtual force models such that it minimizes the deviation from strict rigid body contact conditions (Sec. \ref{sec:trajopt}). The net virtual force acting on a free body is the sum of the virtual forces associated with the contact candidates on that body, $n \in \{0,1,\dots,N_{\Gamma}\}$.



\section{Torque-Limited Planning With Contact}
\label{sec:topus}
\begin{figure*}[h!]
\centering
\includegraphics[width=\textwidth]{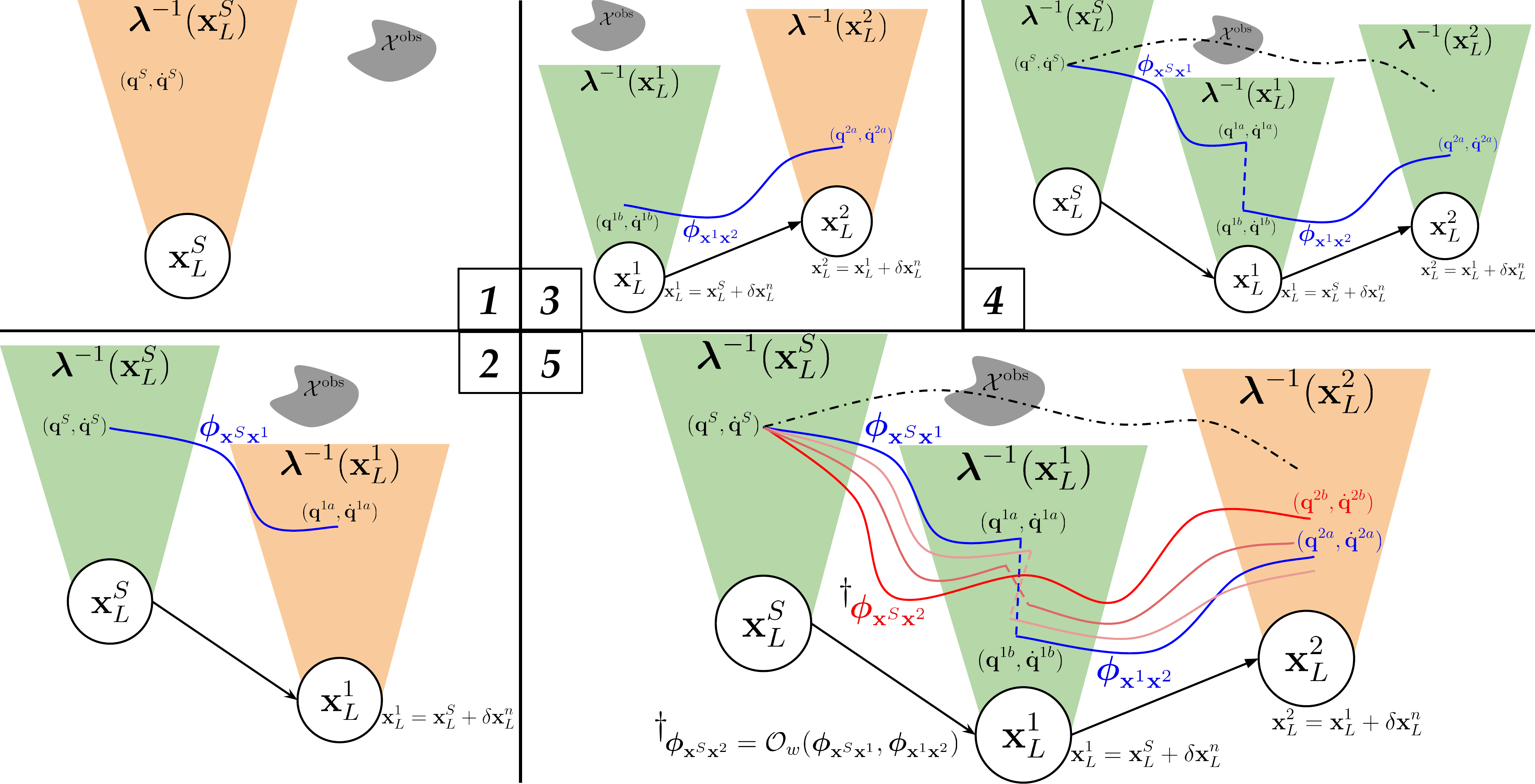}
\caption{Illustration of low-D graph, full-D subspaces of low-D states, trajectory optimization $\mathcal{O}(.)$, warm-started trajectory optimization $\mathcal{O}_w(.)$ and iterating over low-D ancestors (line \ref{linetrajopt:wpsearch}). The illustration above describes an intermediate iteration of the algorithm. The low-D state, the full-D space, and an obstacle in the environment are shown in Box. 1. When a low-D expansion is made (Box. 2), trajectory optimization is used to generate the dynamically feasible trajectory shown in blue that connects the respective full-D spaces. Similarly in Box. 3, the expansion of $\textbf{x}^1_L$ and subsequent lifting to full-D is shown. Following every low-D expansion and subsequent lifting to full-D to find the incremental trajectory from the node being expanded to the successor, the incoming trajectory from the expanded node and outgoing incremental trajectory are warm-started to find the full trajectory to the successor (Box 4 and 5).}
\label{fig:lowdhighd}
\end{figure*}

Our planning framework interleaves graph search with trajectory optimization to combine the benefits of former's ability to search non-convex spaces and solve combinatorial parts of the problem and the latter's ability to obtain a locally optimal solution not constrained to discretization. We will first describe the graph search set up in the low-D space, and then the trajectory optimization in the full-D space that finds the control input trajectory along with the contact model parameters (Fig. \ref{fig:lowdhighd}). We will then explain how INSAT \cite{insat} is adapted for the application of torque-limited manipulation planning with contact. Finally, we provide experimental evidence (Sec. \ref{sec:results}) that INSAT is superior in terms of the solution quality and the planner's behavior than the naive option of running them in sequence. 


\setlength{\textfloatsep}{4pt}
\begin{algorithm*}
\begin{algorithmic}[1]

\Procedure{Key}{$\textbf{x}_L$} \label{line:key}
\State \textbf{return} $g(\textbf{x}_L) + \epsilon*h(\textbf{x}_L)$ \Comment{Compute the priority value for OPEN}
\EndProcedure

\Procedure{Main}{$\textbf{x}^{S}, \textbf{x}^{G}$}
\State $\textbf{x}^S_L = \boldsymbol{\lambda}(\textbf{x}^S); $ $\forall \textbf{x}_L, g(\textbf{x}_L) = \infty$, $\textbf{x}_L.\text{actual} = $ False; $g(\textbf{x}_L^S) = 0$ \Comment{Get low-D start $\textbf{x}_L^S$ and set initial g-values.}
\color{red}
\State $\theta_{\textbf{x}^S\textbf{x}^G}$ = RRTConnect($\textbf{x}^S_L, \textbf{x}^G_L$) \label{line:lowdseedcomp}\color{black}\Comment{Compute low-D seed path}
\color{red}
\State $g(\textbf{x}_L^i) = g(\textbf{x}_L^{i-1}) + \norm{\textbf{x}_L^{i}-\textbf{x}_L^{i-1}}$ where $i=2,\ldots,|\theta_{\textbf{x}^S\textbf{x}^G}|-1$ \color{black}\Comment{Note that $\textbf{x}_L^1 = \textbf{x}_L^S$ and $\textbf{x}_L^{|\theta_{\textbf{x}^S\textbf{x}^G}|} = \textbf{x}_L^G$} 
\color{red}
\State $\forall \textbf{x}_L \in \theta_{\textbf{x}^S\textbf{x}^G};$ Insert $\textbf{x}_L$ in OPEN with \textproc{Key}($\textbf{x}_L$) \label{line:init} \label{line:lowdseedinit} \color{black}\Comment{Insert low-D seed into OPEN list}
\color{black}
\State \textbf{while} \textproc{Key}($\textbf{x}_L^{G}$) $= \infty$ \textbf{do} \label{line:term} \Comment{$\textbf{x}^G_L = \boldsymbol{\lambda}(\textbf{x}^G)$}
\State $\>$ $\>$ $\textbf{x}_L =$ OPEN.$pop()$ \label{line:pq} \Comment{Select the least cost node from OPEN}
\color{magenta}
\State $\>$ $\>$ \textbf{if not} $\textbf{x}_L.\text{actual}$ \textbf{then} \color{black}\Comment{Check whether the lazy cost or actual cost is computed}\color{magenta} \label{line:ifactual}
\State $\>$ $\>$ $\>$ $\>$ $\textbf{x}^{\prime\prime}_L = \textbf{x}_L.\text{pred}$; $\phi_{\textbf{x}^{S}\textbf{x}^{\prime\prime}} = \textbf{x}^{\prime\prime}_L.\text{traj}$; $\phi_{\textbf{x}^{{\prime\prime}}\textbf{x}} = \textbf{x}_L.\text{traj}$ \color{black}\Comment{Get the incoming trajectory to $\textbf{x}_L$ its predecessor}\color{magenta}
\State $\>$ $\>$ $\>$ $\>$ $\phi_{\textbf{x}^{S}\textbf{x}}$ = $\mathcal{O}_w(\phi_{\textbf{x}^{S}\textbf{x}^{\prime\prime}}, \phi_{\textbf{x}^{\prime\prime}\textbf{x}})$ \label{line:trajoptwarm} \color{black}\Comment{Eq. \ref{eq:trajopt} with warm-start. Delayed (lazy) operation postponed during expansion}
\color{orange}
\State $\>$ $\>$ $\>$ $\>$ \textbf{if} $\phi_{\textbf{x}^S\textbf{x}}(T) \notin \lambda^{-1}(\textbf{x}_L)$ \textbf{then} \color{black}\Comment{When the goal constraint is not satisfied in the Boundary Value Problem (BVP)}\color{orange} \label{line:checkreuse}
\State $\>$ $\>$ $\>$ $\>$ $\>$ $\>$ $\textbf{x}_L = \lambda(\phi_{\textbf{x}^S\textbf{x}}(T))$ \color{black}\Comment{Reuse the BVP solution by creating a new state in the graph}
\color{magenta}
\State $\>$ $\>$ $\>$ $\>$ $\textbf{x}_L.\text{traj} = \phi_{\textbf{x}^S\textbf{x}}$ \color{black}\Comment{Update the lazy trajectory with the full trajectory from $\textbf{x}_L^S$}\color{magenta}
\State $\>$ $\>$ $\>$ $\>$ $g(\textbf{x}_L) = J_{total}(\phi_{\textbf{x}^S\textbf{x}})$ \color{black}\Comment{Set the actual cost as the new $g(\textbf{x}_L^S)$}\color{magenta}
\State $\>$ $\>$ $\>$ $\>$ $\textbf{x}_L.\text{actual} = $ True \color{black}\Comment{Set the flag to reflect that the actual cost if found}\color{magenta}
\State $\>$ $\>$ $\>$ $\>$ \textbf{if} $g(\textbf{x}_L) + h(\textbf{x}_L) > $ OPEN.$min()$ \color{black}\Comment{Check if $\textbf{x}_L$ is still OPEN.$min()$ after computing the actual cost}\color{magenta} \label{line:checktoreinsert}
\State $\>$ $\>$ $\>$ $\>$ $\>$ $\>$ Re-insert $\textbf{x}_L$ in OPEN with $\textproc{Key}(\textbf{x}_L)$ \color{black}\Comment{Re-insert back into OPEN if OPEN.$min()$ is different than $\textbf{x}_L$}\color{magenta}
\State $\>$ $\>$ $\>$ $\>$ $\>$ $\>$ \textbf{continue}
\color{black}
\color{red}
\State $\>$ $\>$ $\textbf{X}_L^\prime = \{Succ(\textbf{x}_L) \cup \{\textbf{y}_L \in \theta_{\textbf{x}^S\textbf{x}^G} \mid (\textbf{x}_L, \textbf{y}_L) \in (\mathcal{X}\setminus\mathcal{X}^{obs})\cup\mathcal{X}^S \} \}$ \color{black}\Comment{Check if $\textbf{y}_L \in \theta_{\textbf{x}^S\textbf{x}^G}$ can be a successor of $\textbf{x}_L$}
\color{black} \label{line:succ}
\State $\>$ $\>$ \textbf{for} $\textbf{x}_L^{\prime\prime} \in \textbf{X}_L^\prime$ \textbf{do}  \Comment{Loop over the successors}
\State $\>$ $\>$ $\>$ $\>$ $\textbf{x}_L^\prime = SoftCopy(\textbf{x}_L^{\prime\prime})$\label{line:softcopy} \Comment{State duplicate detection}
\State $\>$ $\>$ $\>$ $\>$ \textbf{if} $\textbf{x}_L^\prime \in $ CLOSED \textbf{then} \label{line:closed}
\State $\>$ $\>$ $\>$ $\>$ $\>$ $\>$ $\textbf{x}_L^\prime = DeepCopy(\textbf{x}_L^{\prime\prime})$  \Comment{If the successor is in CLOSED, then create an augmented state to allow revisiting}
\State $\>$ $\>$ $\>$ $\>$ $\textbf{x}^{\prime\prime}_L, \phi_{\textbf{x}^{\prime\prime}\textbf{x}^\prime}$ = \textproc{GenerateTrajectory}($\textbf{x}_L$, $\textbf{x}_L^\prime$)  \Comment{See Alg. \ref{alg:trajopt}. Only perform incremental optimization and postpone the warm-start optimization to line \ref{line:trajoptwarm}}
\State $\>$ $\>$ $\>$ $\>$ \textbf{if not} $\textbf{x}_L^\prime.\text{actual}$ \textbf{and}  $g(\textbf{x}^{\prime\prime}_L) + J_{total}(\phi_{\textbf{x}^{\prime\prime}\textbf{x}^\prime}) < g(\textbf{x}_L^\prime)$ \textbf{then} \Comment{Check if a better solution is found} \label{line:lazycost}
\State $\>$ $\>$ $\>$ $\>$ $\>$ $\>$ $\textbf{x}_L^{\prime}.\text{pred} = \textbf{x}^{\prime\prime}_L; \textbf{x}^{\prime}_L.\text{traj} = \phi_{\textbf{x}^{\prime\prime}\textbf{x}^\prime}$ \Comment{Update the predecessor and the incoming trajectory}
\State $\>$ $\>$ $\>$ $\>$ $\>$ $\>$ $\textbf{x}_L^\prime.\text{actual} =$ False \Comment{Set the flag to reflect that only the lazy cost is found}
\State $\>$ $\>$ $\>$ $\>$ $\>$ $\>$ \textbf{if} $\textbf{x}_L^{\prime\prime} ==\textbf{x}^S_L$ \textbf{then}
\State $\>$ $\>$ $\>$ $\>$ $\>$ $\>$ $\>$ $\>$ $\textbf{x}_L^\prime.\text{actual} =$ True \Comment{If the predecessor is $\textbf{x}_L^S$, then lazy solution is same as the actual solution.}
\State $\>$ $\>$ $\>$ $\>$ $\>$ $\>$ $g(\textbf{x}_L^\prime) = g(\textbf{x}^{\prime\prime}_L) + J_{total}(\phi_{\textbf{x}^{\prime\prime}\textbf{x}^\prime})$ \Comment{Update the g-value with the better solution}
\State $\>$ $\>$ $\>$ $\>$ $\>$ $\>$ Insert $\textbf{x}_L^\prime$ in OPEN with \textproc{Key}($\textbf{x}_L^\prime$) \Comment{Add the node to OPEN for future expansion}
\EndProcedure
\end{algorithmic}
\caption{INSAT with Lazy Edge Evaluation and Reduced Optimization Rejection}
\label{alg:lazyinsat}
\end{algorithm*}

\subsection{Low-Dimensional Graph Search}

To plan a trajectory that respects system dynamics and controller saturation while simultaneously reasoning globally over large non-convex environments, it is imperative to maintain the combinatorial graph search tractable. To this end, we consider a low-dimensional space $\mathcal{X}_L$ (N-D) comprising joint angles $\textbf{q}$. We build the low-D graph $\mathcal{G}_L$ by discretizing the free joint configuration space of the manipulator $(\mathcal{X} \setminus \mathcal{X}^{\text{obs}}) \cup \mathcal{X}^S$. Each edge in the graph corresponds to the robot's unit joint movement by a known distance (Fig. \ref{fig:lowdhighd}). Every newly generated node is checked to ensure it does not violate joint angle limits and joint torque limits by calculating the gravity compensation before adding it to the graph. For an N DoF manipulator, the branching factor of the graph is $2N$ (unit joint movement in either direction, satisfying joint angle and static joint torque limits). The graph search can be sped up using a heuristic $h(\textbf{x}_L)$, an underestimate of the cost-to-goal of the optimal trajectory. We use the Euclidean distance between two nodes in the joint configuration space as our heuristic, $h(\textbf{x}_L) = \norm{\textbf{x}_L-\textbf{x}_L^G}$.

\subsubsection{Contact vs. Collision} 
\label{sec:cc}
\begin{figure}[h]
\centering
\includegraphics[width=0.7\columnwidth]{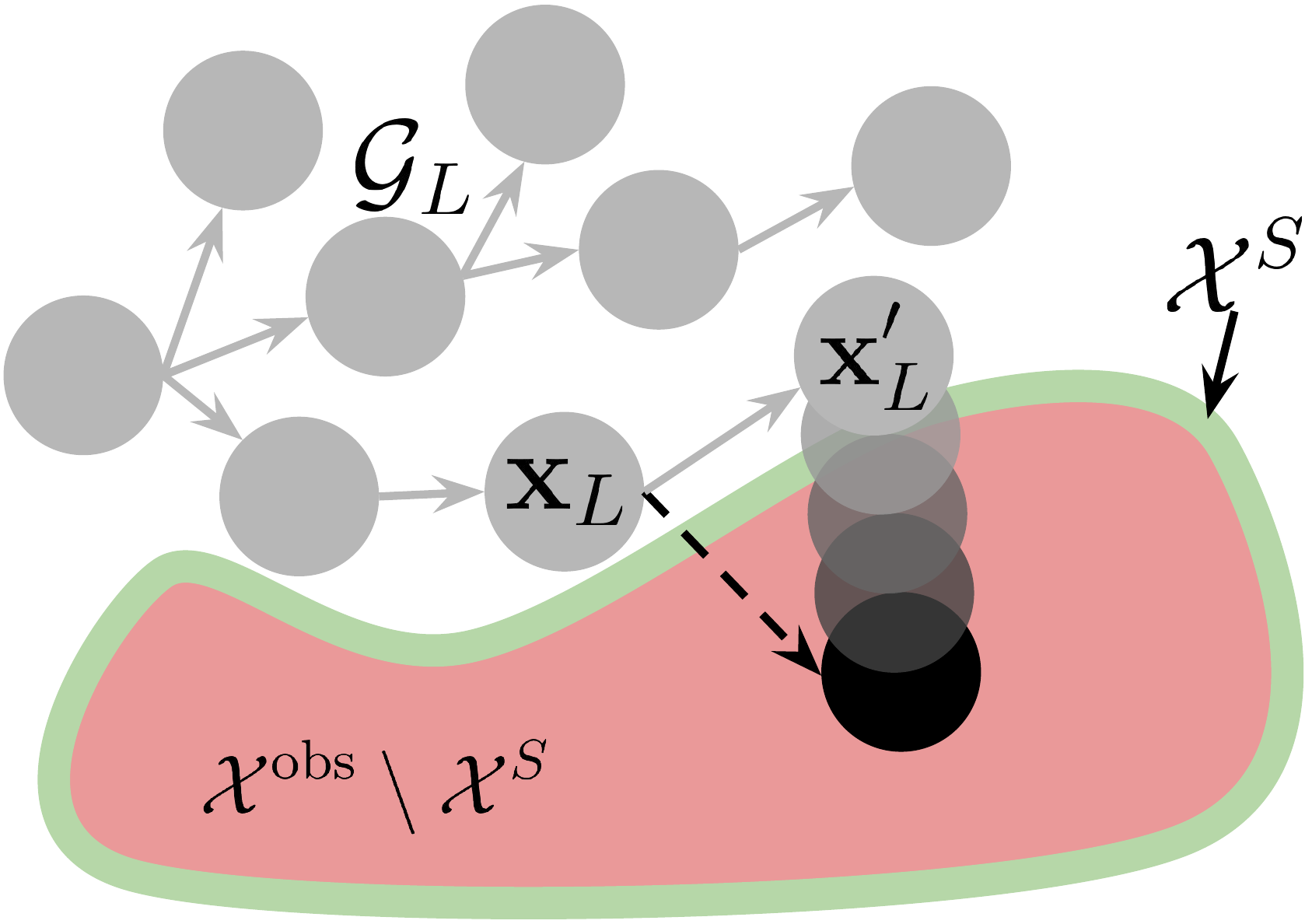}
\caption{Generation of contact configurations. When the low-D state $\textbf{x}_L$ is expanded, the newly generated state is in collision (shown in black with a dashed edge) with $\mathcal{X}^{obs}$. In this work, we use the inbuilt property of MuJoCo to naturally repel the objects in collision to generate the first configuration $\textbf{x}_L^\prime$ that exits from $\mathcal{X}^{\text{obs}} \setminus \mathcal{X}^S$ to $\mathcal{X}^S$ with a contact force gradient under a threshold as a successor.}
\label{fig:contvscoll}

\end{figure}

In motion planning, the task is to find a collision-free path from start to goal. This means making contact or touching the obstacle is considered as collision with the environment. However, for planning with contact, the planner should be allowed to collide (or make contact) with the environment to leverage contact forces and offset robot's limits. To that end, we distinguish contact from collision by defining an obstacle surface $\mathcal{X}^S$. The obstacle surface is a subspace of the obstacle space such that the distance from any point in the obstacle to the free space is bounded by $\beta\to 0_+$.
\begin{equation}
\mathcal{X}^S = \{\textbf{x}\in\mathcal{X}^{\text{obs}} \mid \norm{\textbf{x}-\textbf{x}^{\text{free}}} < \beta , \ \textbf{x}^{\text{free}}\in(\mathcal{X} \setminus \mathcal{X}^{\text{obs}})\}
\end{equation}
When generating successors in the low-D graph (Fig. \ref{fig:contvscoll}), the newly generated successor that is in collision is projected out of the obstacle space by using the intrinsic property of MuJoCo to repel intersecting rigid bodies to generate a state that first exits $\mathcal{X}^{\text{obs}}\setminus\mathcal{X}^S$ with a contact force gradient under a predefined threshold. Such a state typically lies in $\mathcal{X}^S$ and forms the contact configuration.

\subsection{Trajectory Optimization for Planning through Contact}
\label{sec:trajopt}
\subsubsection{Successive Convexification (SCvx)} 
The trajectory optimizer is set up to solve a boundary value problem by finding a joint torque input trajectory that connects the full-D subspaces $\boldsymbol{\lambda}^{-1}(\textbf{x}_L^\prime)$ and $\boldsymbol{\lambda}^{-1}(\textbf{x}_L^{\prime\prime})$ of two manipulator configurations $\textbf{x}_L^\prime$ and $\textbf{x}^{\prime\prime}_L$. In the conference version of the paper, we used Successive Convexification (SCvx) to solve the boundary value problem\cite{citoonol}. SCvx solves a sequence of smooth quadratic approximations of the original nonlinear problem subjected to linearized dynamics. But, as the manipulator dynamics with contact is discontinuous, the linearization of dynamics is poor. To alleviate this, we use the tunable soft contact model (Sec \ref{sec:vscm}) to solve the trajectory optimization problem (Eq. \ref{eq:trajopt}). We begin with the relaxed setting for the contact model (\textit{i.e.} large values for $\textbf{k}= [k_1, k_2,\dots, k_{N_\Gamma}]\intercal, \textbf{b}= [b_1, b_2,\dots, b_{N_\Gamma}]\intercal, \boldsymbol{\mu}= [\mu_1, \mu_2,\dots, \mu_{N_\Gamma}]\intercal$ that correspond to non-zero virtual contact forces when not in contact and nonzero virtual frictional force when the object is at rest) in which the system dynamics (manipulator dynamics + contact dynamics) and its gradients are smooth and solve the Eq. \ref{eq:trajopt} using SCvx.   
\begin{subequations}
\begin{align}
\begin{split}
\min_{\textbf{u}[.], \textbf{k}, \textbf{b}, \boldsymbol{\mu}}\qquad & l(\textbf{x}, \textbf{u}) =
  w_1\norm{\textbf{x}_L[N] - \textbf{x}^{G}_L} + \\ \sum_{i=0}^{N-1} (w_2\norm{\textbf{u}[i]} +& w_3\norm{\dot{\textbf{x}}[i]}) + \sum_{n=1}^{N_{\Gamma}} w_4\norm{\textbf{k}} + w_5\norm{\textbf{b}} + w_6\norm{\boldsymbol{\mu}}
\end{split}
\label{eq:trajoptobj} 
\\
\text{s.t.}\qquad & \textbf{x}[0] = \textbf{x}^0; \textbf{x}[i+1] = \textbf{f}(\textbf{x}_i, \textbf{u}_i)\label{eq:constraint1} 
\\
& |\dot{\textbf{x}}(t)| \leq \dot{\textbf{x}}_{\text{lim}}; |\ddot{\textbf{x}}(t)| \leq \ddot{\textbf{x}}_{\text{lim}}; |\textbf{u}(t)| \leq \textbf{u}_{\text{lim}}
\label{eq:constraint2}
\end{align}
\label{eq:trajopt}
\end{subequations}
where $w_1,\ldots, w_6$ represent the relative importance of the cost terms. The virtual forces disappear when $\textbf{k}, \textbf{b}, \boldsymbol{\mu} = \textbf{0}$ (Eq. \ref{eq:normcontmod}, \ref{eq:normtofric}) and the above minimization problem optimizes for that. Although Eq. \ref{eq:trajopt} looks like direct shooting, the variant of SCvx combines the benefit of shooting and direct transcription by exploiting the sparsity in linear dynamics constraint during the convexification phase and maintaining dynamic consistency by rolling out the trajectory with the full nonlinear dynamics using MuJoCo (see \cite{citoonol}). 

Note that SCvx is very similar to indirect methods like differential dynamic programming (DDP) or its first-order simplification, iterative linear quadratic regulator (iLQR), with the difference that SCvx computes the backward pass updates by maintaining a trust region. The trust region is in turn grown or shrunk depending on the quality of the update to the decision variables and forward pass divergence. This means that the single update step in SCvx in the best case is the same as iLQR and will be slower than iLQR in the cases when the trust region has to be modified. In SCvx, despite using smooth tunable contact models to prevent abrupt changes to the cost function derivatives, solving the optimization reliably required careful tuning of the step size of the simulation and computing the derivatives. In other words, the optimization formulation was sensitive near contact configurations. Based on this observation, we updated the cost function using a classical risk-sensitive formulation \cite{mjpc}. This dramatically improved the reliability of the optimization convergence and reduced the number of times the trust region had to be modified.

\begin{algorithm}
\begin{algorithmic}[1]

\Procedure{GenerateTrajectory}{$\textbf{x}_L$, $\textbf{x}_L^\prime$}
\State \textbf{for} $\textbf{x}_L^{\prime\prime} \in Ancestors(\textbf{x}_L)$ \textbf{do} \label{linetrajopt:wpsearch} \Comment{From $\textbf{x}_L$ to $\textbf{x}^S_L$}
\State $\>$ \textbf{if} \ $\textbf{x}_L^{\prime\prime} = \textbf{x}^S_L$ \textbf{then}
\State $\>$ $\>$ $\boldsymbol{\lambda}^{-1}(\textbf{x}_L^{\prime\prime}) = \textbf{x}^S$
\State $\>$ \textbf{else if} $\textbf{x}_L^\prime = \textbf{x}^G_L$ \textbf{then}
\State $\>$ $\>$ $\boldsymbol{\lambda}^{-1}(\textbf{x}_L^{\prime}) = \textbf{x}^G$
\State $\>$ $\phi_{\textbf{x}^{\prime\prime}\textbf{x}^\prime}$ = $\mathcal{O}(\boldsymbol{\lambda}^{-1}(\textbf{x}_L^{\prime\prime}), \boldsymbol{\lambda}^{-1}(\textbf{x}_L^\prime))$ \Comment{Sec. \ref{sec:trajopt}, Eq. \ref{eq:trajopt}}\label{linetrajopt:trajoptgen}
\State $\>$ \textbf{if} $\phi_{\textbf{x}^{\prime\prime}\textbf{x}^\prime}.IsCollisionFree()$ \textbf{then} \Comment{Sec. \ref{sec:cc}} \label{linetrajopt:collfeas}
\State $\>$ $\>$ \textbf{return} \{$\textbf{x}^{\prime\prime}_L, \phi_{\textbf{x}^{\prime\prime}\textbf{x}^\prime}$\} \label{linetrajopt:retbasicphase}
\State \textbf{return} NULL w/ discrete $\infty$ cost
\EndProcedure
\end{algorithmic}
\caption{Trajectory Optimization for INSAT}
\label{alg:trajopt}
\end{algorithm}

We will now present the first upgrade to our previous work \cite{insat_ptc}. The next section will explain the other three upgrades. 
\subsubsection{Risk-sensitive Formulation}
The cost function of the optimization in Eq. \ref{eq:trajopt} takes the form
\begin{equation}
    l(\textbf{x}, \textbf{u}) = \sum_{i=0}^{M} w_i \norm{\textbf{r}_i(\textbf{x}, \textbf{u})}  
    \label{eq:riskbasecost}
\end{equation}

The above cost is expressed as a sum of $M$ terms, each made of three components. Firstly, a non-negative weight $w_i \in \mathbb{R}_+$ is assigned to signify the relative significance of the particular term within the cost function. Secondly, the norm function $\norm{.}: \mathbb{R}^p \rightarrow \mathbb{R}_+$ takes in a vector and returns a non-negative scalar and achieves minimum at \textbf{0}. This norm function is crucial in characterizing the behavior of the involved variables. Lastly, the residual $r \in \mathbb{R}$ is introduced as a vector of elements that tend to be small when the underlying task is accomplished. Together, these components contribute to the comprehensive formulation of the cost function, capturing both the importance weights and the characteristics of the residual vector in the optimization process. We augment the cost function, as represented by Eq. \ref{eq:riskbasecost}, by incorporating a risk-aware exponential scalar transformation denoted as $\xi : \mathbb{R}_+ \times \mathbb{R} \rightarrow \mathbb{R}$. This transformation aligns with the principles of classical risk-sensitive control, as elucidated in \cite{mjpc}. The resultant running cost $c$ is given by 

\begin{equation}
    c(\textbf{x}, \textbf{u}) = \xi(l(\textbf{x}, \textbf{u}), R) = \frac{e^{R. l(\textbf{x}, \textbf{u})} - 1}{R}
    \label{eq:riskexpcost}
\end{equation}

The scalar parameter $R \in \mathbb{R}$ serves as an indicator of risk-sensitivity, with $R=0$ (the default) denoting risk-neutrality, $R>0$ reflecting risk-aversion, and $R<0$ indicating risk-seeking behavior. The mapping function $\xi$ exhibits the following characteristics: it is well-defined and smooth for any value of $R$; when $R=0$, it reduces to the identity function $\xi(l, 0) = l$ in the limit; zero is a fixed point, as $\xi(0, R) = 0$; the derivative of $\xi$ with respect to $l$ at 0 is 1, expressed as $\partial \xi(0, R)/\partial l = 1$; for $R<0$, $\xi$ is bounded, specifically $\xi(l, R) = -1/R$; the output of $\xi(l, R)$ shares the same units as $l$; it is monotonic, with $\xi(l, R)$ greater than $\xi(z, R)$ if $l > z$; and non-negative, ensuring $\xi(l, R) > 0$ when $l > 0$. 

\textbf{Gradients:} The gradients of the cost function can be derived as follows

\begin{equation}
\begin{aligned}
& \frac{\partial c}{\partial \textbf{x}}=e^{R l} \frac{\partial l}{\partial \textbf{x}}=e^{R l} \sum_{i=0}^M w_i \frac{\textbf{r}_i}{\norm{\textbf{r}_i}} \frac{\partial \textbf{r}_i}{\partial \textbf{x}} \\
& \frac{\partial c}{\partial \textbf{u}}=e^{R l} \frac{\partial l}{\partial \textbf{u}}=e^{R l} \sum_{i=0}^M w_i \frac{\textbf{r}_i}{\norm{\textbf{r}_i}} \frac{\partial \textbf{r}_i}{\partial \textbf{u}}
\end{aligned}
\label{eq:grad}
\end{equation}


\textbf{Hessians:} Similarly the second-order derivatives can be derived using the Gauss-Newton approximation as

\begin{equation}
\begin{gathered}
\frac{\partial^2 c}{\partial \textbf{x}^2} \approx e^{R l}\left[\sum_{i=0}^M w_i \frac{\partial \textbf{r}_i^T}{\partial \textbf{x}} \left( \frac{1}{\norm{\textbf{r}_i}}-\frac{\textbf{r}_i^2}{\norm{\textbf{r}_i}^3} \right) \frac{\partial \textbf{r}_i}{\partial \textbf{x}}+R \frac{\partial l^T}{\partial \textbf{x}} \frac{\partial l}{\partial \textbf{x}}\right] \\
\frac{\partial^2 c}{\partial \textbf{u}^2} \approx e^{R l}\left[\sum_{i=0}^M w_i \frac{\partial \textbf{r}_i^T}{\partial \textbf{u}} \left( \frac{1}{\norm{\textbf{r}_i}}-\frac{\textbf{r}_i^2}{\norm{\textbf{r}_i}^3} \right) \frac{\partial \textbf{r}_i}{\partial \textbf{u}}+R \frac{\partial l^T}{\partial \textbf{u}} \frac{\partial l}{\partial \textbf{u}}\right] \\
\frac{\partial^2 c}{\partial \textbf{x} \partial \textbf{u}} \approx e^{R l}\left[\sum_{i=0}^M w_i \frac{\partial \textbf{r}_i^T}{\partial \textbf{x}} \left( \frac{1}{\norm{\textbf{r}_i}}-\frac{\textbf{r}_i^2}{\norm{\textbf{r}_i}^3} \right) \frac{\partial \textbf{r}_i}{\partial \textbf{u}}+R \frac{\partial l}{\partial \textbf{x}} \frac{\partial l}{\partial \textbf{u}}\right]
\end{gathered}
\label{eq:hess}
\end{equation}
where the second-order derivatives of $\textbf{r}$ are ignored.


\subsubsection{Iterative Linear Quadratic Regulator (iLQR)} The aforementioned risk-sensitive transformation (Eq. \ref{eq:riskexpcost}) is applied to $l(\textbf{x}, \textbf{u})$ in Eq. \ref{eq:trajoptobj}. The resulting overall cost function $c(\textbf{x}, \textbf{u})$ is used for running the iLQR. The iLQR is a Gauss-Newton approximation of the DDP algorithm \cite{ddp}, utilizes first and second-order derivative information (Eq. \ref{eq:grad} and \ref{eq:hess}) to take an approximate Newton step over the open-loop control sequence $\textbf{u}_{0:T}$ via dynamic programming, producing a time-varying linear feedback policy

\begin{align}
    \textbf{u} = \bar{\textbf{u}} + \textbf{K} (\textbf{x} - \bar{\textbf{x}}) + \alpha \textbf{k}
\end{align}
The nominal or current optimal trajectory is indicated by overbars ( $\bar{}$ ), where $\textbf{K}$ represents the feedback gain matrix, and $\textbf{k}$ denotes an update to the current action trajectory. A parallel line search over the step size $\alpha \in [\alpha_{min}, 1]$ is conducted to identify the most favorable improvement. Further enhancements include a constrained backward pass \cite{ddp}, enforcing action limits and incorporating adaptive regularization. For a detailed explanation of iLQR/DDP, we ask the reader to refer to \cite{ilqg, ddp}. 

The action trajectory $\textbf{u}(t)$ is represented as a spline to reduce the search space. Splines are part of a class of linear bases. In contrast to other linear bases, splines possess a useful characteristic: constraining the spline parameter values automatically constrains the spline trajectory. This holds precisely true for zero and linear interpolations and is largely accurate for cubic splines. The significance of bounding arises from the common practice in most physical systems where controls are limited to specific bounds, making it unnecessary to search beyond these established limits.


\subsection{INSAT: INterleaved Search And Trajectory Optimization}

\begin{figure}[h]
\includegraphics[width=\columnwidth]{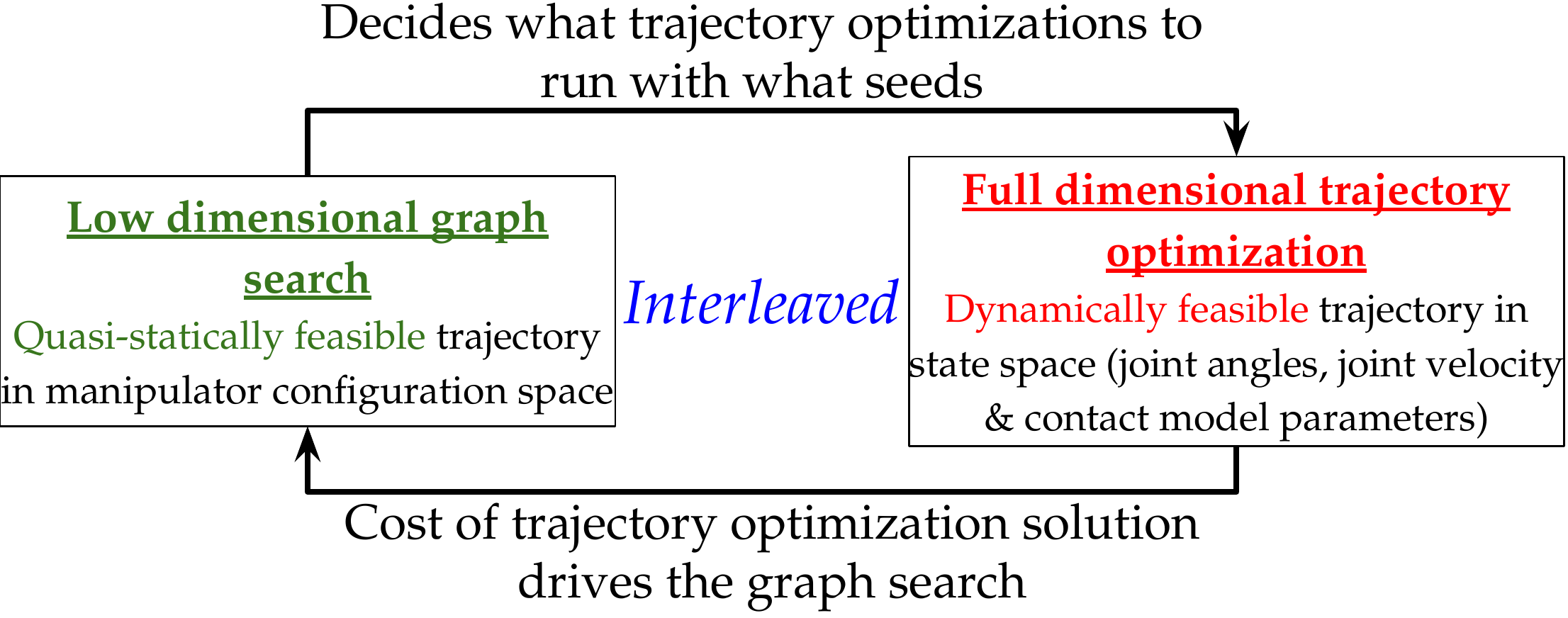}
\caption{A schematic of the working principle of INSAT}
\label{fig:insat_flow}
\end{figure}



\begin{figure*}[h!]
    \center
    \includegraphics[width=\textwidth]{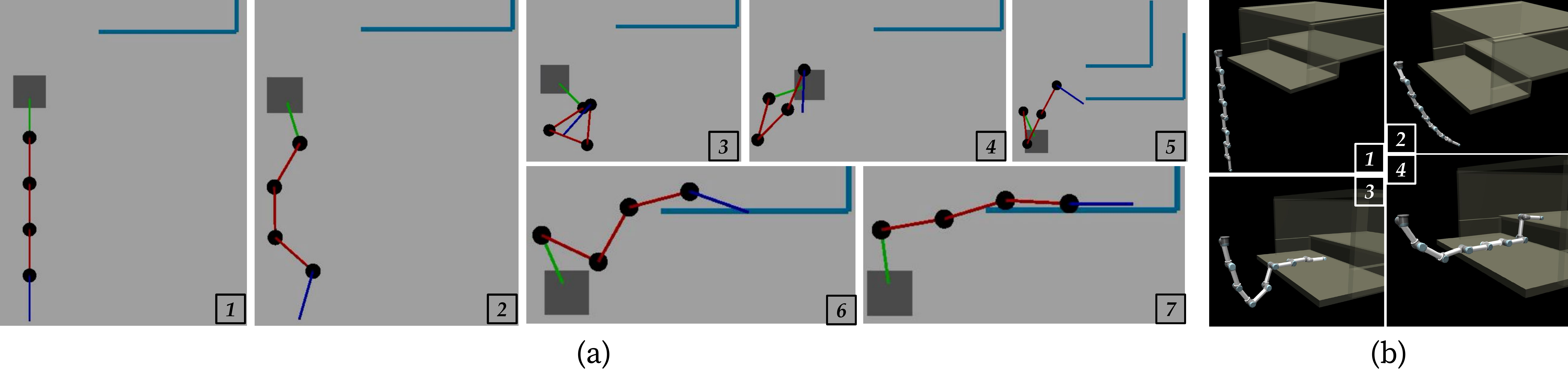}
    \caption{Simulation experiments on (a) a planar arm, (b) hyper-redundant arm and (c) Kinova Gen3. In (a), the planar arm swings its way up into the ledge. It first rolls itself into a compact configuration (a3) to minimize the net torque by minimizing the moment arm and unfolds with the accrued momentum (a4) into the ledge. Once on the ledge, it slides its way by staying in contact and expending least joint effort. In (b), similar behavior as (a) is exhibited but on a 9 DoF redundant system. The robot has to slide and climb the shelf in a stretched out configuration. The robot climbs the stair without breaking contact.}
    \label{fig:swinga}
\end{figure*}

\begin{figure*}[h!]
    \center
    \includegraphics[width=\textwidth]{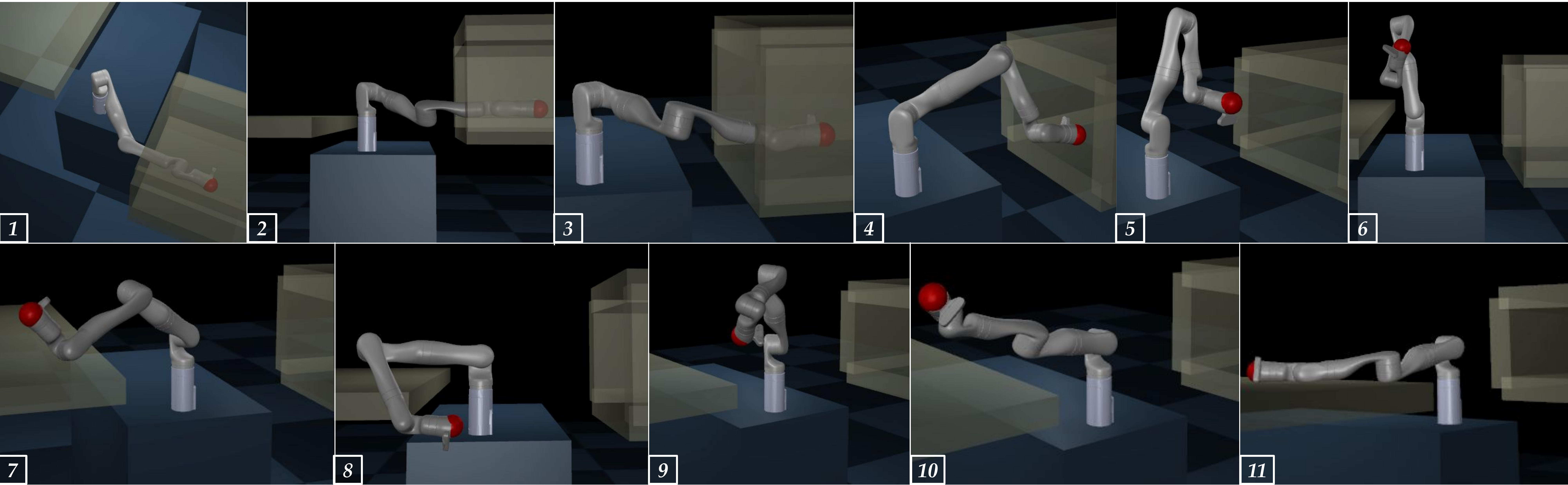}
    \caption{The Kinova Gen3 robot picks and places an overweight object (shown in red) from a confined shelf to a table. The robot's payload limit is 4kg and we used a 4.7kg payload. The mass violates the static torque limits without contact support from the environment at the start configuration. The robot maintains the contact with the shelf as much as possible by dragging the object out and swinging across its base to pump energy to eventually carry the object on to the table. The task requires reasonably long horizon planning in which standalone trajectory optimization struggles. By guiding the trajectory optimization with graph search over manipulator configurations, INSAT is able to produce a dynamically feasible trajectory of unique behavior.}
    \label{fig:swingb}
\end{figure*}

An overview of the algorithm is presented using a flowchart in Fig. \ref{fig:insat_flow}. INSAT performs interleaved search on discrete low-dimensional manipulator configuration space and continuous high-dimensional joint velocity and contact model parameter space. The low dimensional search gets the manipulator around obstacles and evaluates various contact mode sequences. The high-dimensional trajectory optimization validates or invalidates the dynamic feasibility of paths discovered by the low-dimensional search. Consequently, INSAT generates dynamically feasible trajectories for the manipulator to brace with the environment, offset/stay within its torque limits, and reach the desired goal.

\begin{figure*}[h]
\centering
\includegraphics[width=\textwidth]{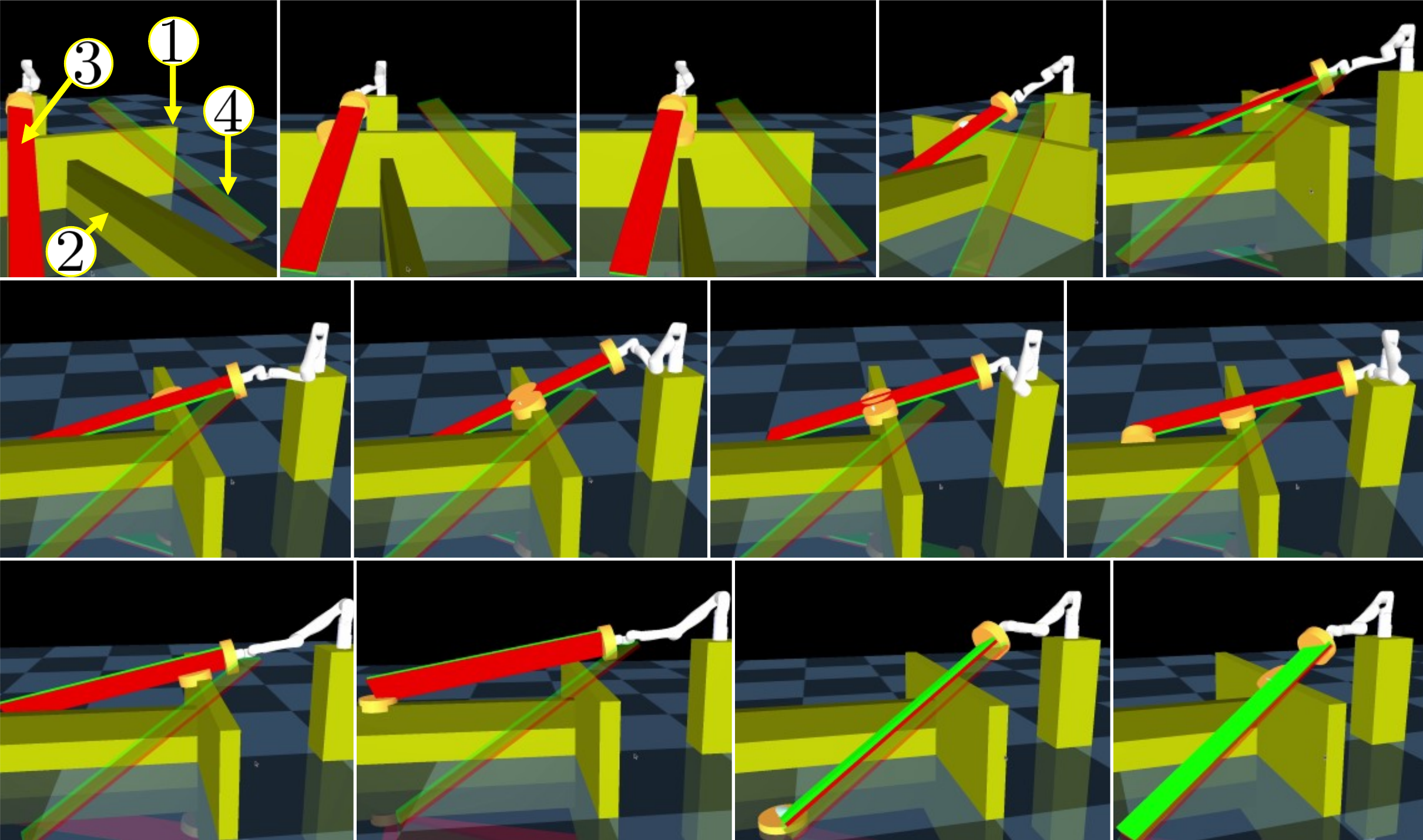}
\caption{Filmstrip showing the Kinova Gen3 arm flipping a 3kg payload across a medium difficulty partition in simulation. The contact locations of the payload with the arm and partition wall are depicted with orange cylinders. ROW 1: The arm begins by dragging the payload towards the partition wall and slides the payload out towards its base. ROW 2: The arm makes further progress toward the goal by navigating in a very constrained manifold, wiggling the payload to the other side of the partition while maintaining continuous contact with the front wall. ROW 3: Finally, the arm uses the front wall as a fulcrum to support the payload against the partition and drags it into the goal configuration.}
\label{fig:flip_level}
\end{figure*}

Alg. \ref{alg:lazyinsat} presents the pseudocode of INSAT for torque-limited manipulation planning with contact. The algorithm takes as input the full-D start and goal states $\textbf{x}^{S}$ and $\textbf{x}^{G}$. To search in the low-D graph $\mathcal{G}_L$, we use weighted A* (WA*)\cite{pohlwastar} which maintains a priority queue called OPEN that dictates the order of expansion of the states and the termination condition based on \textproc{Key($\textbf{x}_L$)} value (Alg. \ref{alg:lazyinsat}: lines \ref{line:key}, \ref{line:term}). Alg. \ref{alg:lazyinsat} maintains two functions: cost-to-come $g(\textbf{x}_L)$ and a heuristic $h(\textbf{x}_L)$. $g(\textbf{x}_L)$ is the cost of the current path from the $\textbf{x}^S$ to $\textbf{x}_L$ and $h(\textbf{x}_L)$ is an underestimate of the cost of reaching the goal from $\textbf{x}_L$. WA* initializes OPEN with $\textbf{x}_L^{S}$ (Alg. \ref{alg:lazyinsat}: line \ref{line:init}) and tracks the expanded states using another list called CLOSED (Alg. \ref{alg:lazyinsat}: line \ref{line:closed}). 

A graphical illustration of how the low-D state expansions and full-D trajectory generations might look is shown in Fig \ref{fig:lowdhighd}. Each time the search expands a state $\textbf{x}_L$, it removes $\textbf{x}_L$ from OPEN and generates the successors as per the discretization (Alg. \ref{alg:lazyinsat}: lines \ref{line:pq}, \ref{line:succ}). For every low-dimensional successor $\textbf{x}_L^\prime$, we solve a trajectory optimization problem described in Sec. \ref{sec:trajopt} to find a corresponding full-D trajectory from the full-D subspace of the closest ancestor $\lambda^{-1}(\textbf{x}^{\prime\prime}_L)$ of $\textbf{x}_L$ (Alg. \ref{alg:trajopt}: line \ref{linetrajopt:wpsearch}) to $\boldsymbol{\lambda}^{-1}(\textbf{x}_L^\prime)$ (Alg. \ref{alg:trajopt}: line \ref{linetrajopt:trajoptgen}, Fig \ref{fig:lowdhighd}). The trajectory optimization output $\phi_{\textbf{x}^{\prime\prime}\textbf{x}^\prime}$ is checked for collision (Alg. \ref{alg:trajopt}: line \ref{linetrajopt:collfeas}, Sec. \ref{sec:cc}). If the optimized trajectory $\phi_{\textbf{x}^{\prime\prime}\textbf{x}^\prime}$ is in collision or infeasible (Fig \ref{fig:lowdhighd}), the algorithm continues with the next closest ancestor (Alg. \ref{alg:trajopt}: line \ref{linetrajopt:wpsearch}). Upon finding the state $\textbf{x}_L^{\prime\prime}$ which enables a full-dimensional feasible trajectory $\phi_{\textbf{x}^{\prime\prime}\textbf{x}^\prime}$, the entire trajectory from start $\phi_{\textbf{x}^{S}\textbf{x}^\prime}$ is constructed by warm-starting the optimization ($\mathcal{O}_w$) with the trajectories $\phi_{\textbf{x}^{S}\textbf{x}^{\prime\prime}}$ and the newly generated trajectory $\phi_{\textbf{x}^{\prime\prime}\textbf{x}^\prime}$ (Alg. \ref{alg:lazyinsat}: line \ref{line:trajoptwarm}) by relaxing all the waypoint and derivative constraints (Fig. \ref{fig:lowdhighd}) until convergence or trajectory becoming infeasible, whichever occurs first. In \cite{insat_ptc}, the optimization from scratch to find $\phi_{\textbf{x}^{\prime\prime}\textbf{x}^\prime}$ and the warm-starting to find the entire trajectory $\phi_{\textbf{x}^{S}\textbf{x}^{\prime\prime}}$ happens as consecutive steps. Here the second step of warm-starting the entire optimization is the slower of the two. In the following section on Lazy INSAT, we explain how this slower step can delayed and avoided until needed. 

\section{Delaying and Re-purposing Optimization Calls in INSAT}
The following subsections will explain the remaining three upgrades to the conference version of our work \cite{insat_ptc}. The colored lines in Alg. \ref{alg:lazyinsat}. match the color of the title text of the corresponding subsections. 

\subsubsection{\color{magenta}Lazy INSAT}

A comprehensive exposition of INSAT is presented in the previous section and \cite{insat}. In this section, we delve into the intricacies of Lazy INSAT, an adaptation of INSAT that incorporates lazy edge evaluation depicted by magenta lines in the pseudocode Alg. \ref{alg:lazyinsat}. The rationale behind the introduction of Lazy INSAT stems from a critical observation in the realm of indirect trajectory optimization techniques.

Our key insight lies in the realization that warm-starting large problems yields a considerable reduction in computational time compared to solving the same large problem from scratch. However, it is important to note that warm-starting a large problem is not as swift as solving smaller problems from scratch. In the context of Lazy INSAT, the term ``small problems'' refers to the incremental optimizations undertaken to determine the trajectory from the state being expanded to its successor ($\phi_{\textbf{x}^{\prime\prime}\textbf{x}^\prime}$ in Alg. \ref{alg:trajopt}: line \ref{linetrajopt:trajoptgen}). Conversely, "warm-starting" entails establishing the complete trajectory from the initial state to the successor (Alg. \ref{alg:lazyinsat}: line \ref{line:trajoptwarm}). To simplify, while the warm-starting step in the original INSAT significantly enhances runtime and aids in the convergence of long-horizon problems, the efficiency gains diminish as the horizon lengthens. Therefore, we propose postponing the warm-starting optimizations whenever feasible.

The crux of the lazy approach lies in maintaining a lower bound on the solution of trajectory optimization that is fast to compute and treating it as a pseudo cost to facilitate ordering the priority queue (Alg. \ref{alg:lazyinsat}: lines \ref{line:trajoptwarm}-\ref{line:lazycost}). Notably, the actual cost is only computed when a node is selected for expansion, not during its generation as in INSAT \cite{insat_ptc} (Alg. \ref{alg:lazyinsat}: line \ref{line:ifactual}). The true cost contributes to increasing the g-value (cost-so-far) of the node chosen for expansion. This adjustment means that the chosen node need not always be at the top of the OPEN list (Alg. \ref{alg:lazyinsat}, line \ref{line:checktoreinsert}). In cases where the selected node is displaced from the top position, it is reinserted into the OPEN list, and the subsequent state at the top becomes the new focus for expansion. Expansion of the chosen state only occurs when its true cost is the lowest among the costs associated with nodes in the OPEN list. This lazy evaluation strategy aims to strike a balance between computational efficiency and optimization accuracy, ensuring that the costliest computations are deferred until their necessity becomes evident during the expansion process. The overall objective is to enhance the efficiency of INSAT, particularly for scenarios with extended planning horizons.

\subsubsection{\color{orange}Reduced Rejection of Optimized Trajectory}
In INSAT, the graph search operates within a space that is discrete and low-dimensional. This means that the optimization process needs to navigate and connect discrete cell centers, creating a challenge akin to solving a boundary value problem. In practice, to address the complexities of this task, penalty-based optimization is employed, allowing for some relaxation of the strict constraints. While these methods enhance the behavior of the optimization, there are instances where the optimization successfully identifies a feasible trajectory but falls short of satisfying the required boundary conditions. As we use indirect methods for trajectory optimization, they depend on rollouts to generate dynamically feasible trajectories. This inherently ensures the satisfaction of the starting point condition, as it relies on the generation of trajectories from initial conditions. However, the challenge lies in situations where the optimization process yields feasible trajectories that, while dynamically sound, do not fully meet the specified boundary conditions. To tackle this issue, a strategy is employed to reuse these feasible yet non-convergent trajectories. This involves the introduction or update of a new or existing node in the low-dimensional graph. The decision for this update is based on the terminal point of the reused trajectory (Alg. \ref{alg:lazyinsat}, line \ref{line:checkreuse}). By incorporating these trajectories and updating the graph accordingly, the optimization process gains a degree of adaptability. This adaptive strategy helps address scenarios where the optimization converges to a feasible trajectory but struggles with meeting the necessary boundary conditions, contributing to a more robust and versatile optimization process within the INSAT framework (orange lines in Alg. \ref{alg:lazyinsat}).
\subsubsection{\color{red} Seed Low-D search with RRT-Connect} Bidirectional sampling-based planners like RRT-connect can be very fast for high dimensional manipulation problems at the cost of poor solution quality, Moreover, planners like RRT-connect do not produce dynamically feasible trajectories. We use the RRT-connect solution to seed the low-D discrete graph search by inserting the OPEN priority queue with the nodes on the RRT-connect solution at the start of the algorithm (red lines in Alg. \ref{alg:lazyinsat}). The RRT-connect path is retained in a separate data structure and during every expansion in the search, we loop over the RRT-connect path in reverse and check if there is a valid edge from the state being expanded to a node on the RRT-connect path (Alg. \ref{alg:lazyinsat}, line \ref{line:succ}). If a valid edge is found, we add it to the set of successors. If there also exists a valid high-D trajectory to this successor, then it is added to the OPEN list for possible future expansion. Intuitively, these nodes from the RRT-Connect serve as visibility nodes to escape local minima while preserving the bounded suboptimality guarantees of the graph search. 




\section{Experiments and Results}
\label{sec:results}

The adaptation of INSAT for planning with contact proposed in \cite{insat_ptc} was evaluated on four simulation experiments and one hardware experiment. For the enhancements to \cite{insat_ptc} presented here, we add a new simulation experiment and a new hardware experiment. The simulation experiments are carried out using the open-source MuJoCo simulator \cite{mujoco}. For hardware experiments, we use the Kinova Gen3 7-DoF manipulator. We also provide additional analysis to empirically support our choice of low-D search and full-D trajectory optimization algorithm. All the methods methods were implemented in C++ on AMD Ryzen Threadripper Pro.

\begin{figure*}
\begin{subfigure}{\textwidth}
\includegraphics[width=\textwidth]{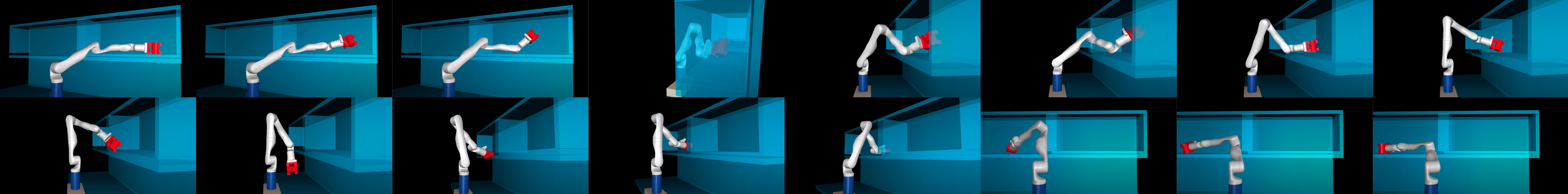}
\caption{Payload transportation between adjacent shelves using contact in simulation.}
\label{fig:shelfsim}
\end{subfigure}
\begin{subfigure}{\textwidth}
\includegraphics[width=\textwidth]{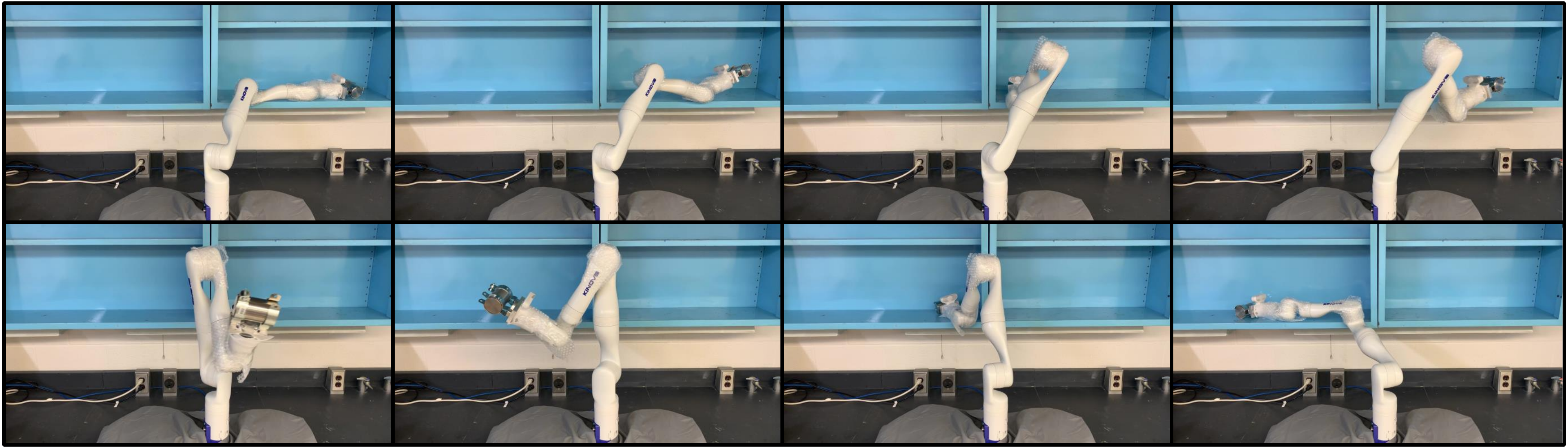}
\caption{Payload transportation between the adjacent shelves using contact on the hardware.}
\label{fig:shelfhw}
\end{subfigure}
\caption{Film strips showing a 7 DoF Kinova Gen3 robot utilizing bracing contacts to transfer a 2.5 kg payload between two cabinets using minimal torque. The arm slides the payload to the center by bracing with its wrist before lifting on its own. The payload is then placed at the proximal end of the target shelf and pushed by bracing its forearm.}
\end{figure*}


\begin{figure}[h]
\centering
\includegraphics[width=\columnwidth]{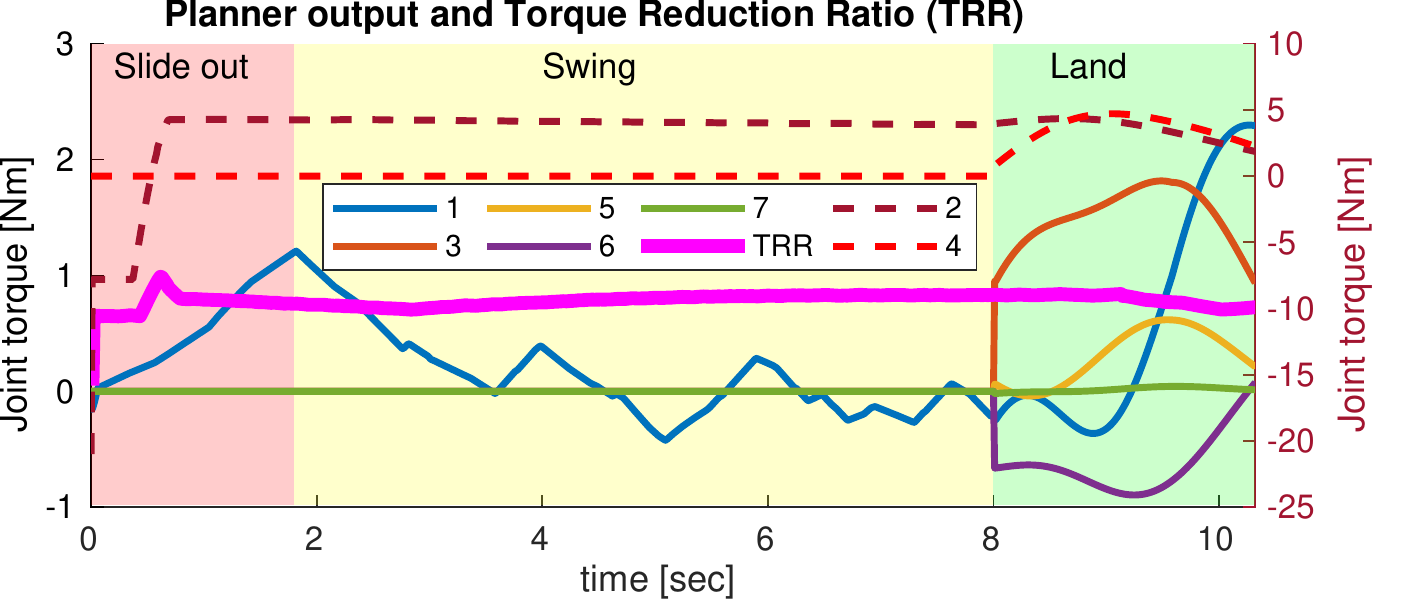}
\caption{Joint torque trajectory for simulation scenario (c) and TRR calculated with respect to our baseline (RRT-Connect + direct collocation). The torque plot for joints 2 and 4 using dotted lines use the y-axis on the right.}
\label{fig:torqsw}
\end{figure}

\begin{figure}[h]
\centering
\includegraphics[width=\columnwidth]{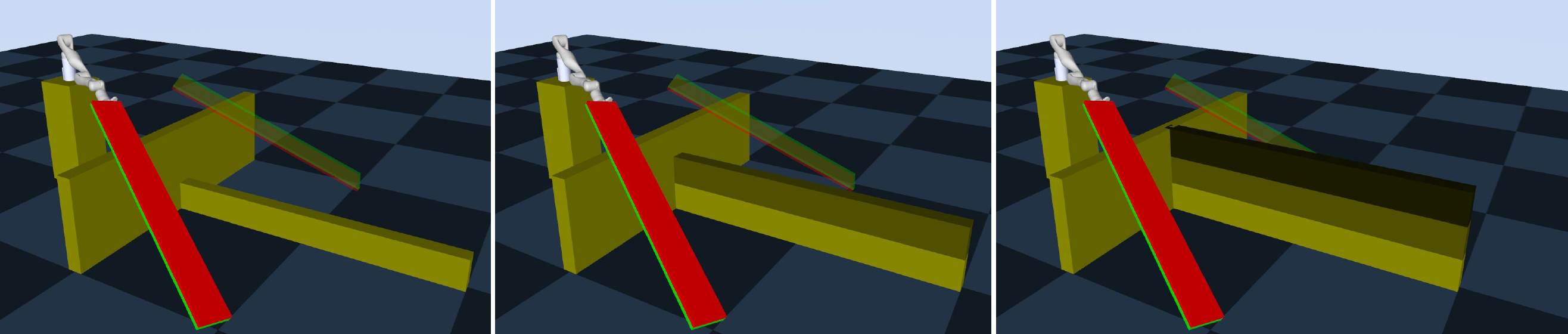}
\caption{Easy, Medium, and Hard difficulty levels based on the height of the central partition for the flipping task. From the left to the right, the height of the partition and the complexity of the task increases.}
\label{fig:flip_level_2} 
\end{figure}


From what we can ascertain, the proposed method is the first algorithm to introduce global, long-horizon manipulation planning through bracing. This implies the ability to plan, reason and execute manipulation actions through contact (bracing) over an extended time frame. As such, we could not find a perfect baseline for comparison that solves our exact problem. The method that comes closest to ours in terms of its ability to solve similar problems is \cite{pangcontact}. By making the quasi-static assumption, \cite{pangcontact} reduces the state space dimension and achieves significant speedup in solving complex contact-rich tasks. In contrast, we generate plans with full dynamic feasibility and combat the high dimensionality of the state space by interleaving searching in low-D space and optimizing in full-D space. 

\textbf{Interleaved vs. Sequential Planning:} We demonstrate the benefits of interleaving graph search and contact implicit trajectory optimization in comparison to the more common approach of using them in sequence \cite{contsearch1,contsearch2,contsearch3,contrrt}. To do so, we compare our method with a sequential combination of RRT-Connect (Rapidly Exploring Random Trees) and direct collocation \cite{contrrt}. The choice of RRT-Connect over discrete graph search algorithms is that RRT variants are a popular choice for fast high-dimensional manipulation planning. 




\subsection{Simulation Experiments}
For the simulation experiments, we consider five different environments and three 
different robot types namely (a) a planar 5-link arm climbing a ledge, (b) a contrived hyper-redundant (9 DoF) version of UR5 manipulator that has to enter a rectangular tube and crawl over a step, (c) a pick and place task of an overweight payload, (d) a payload transportation task between adjacent shelves, and (e) a bulky payload flipping and transportation across a partition wall. The simulations (c), (d), and (e) are carried out on a Kinova Gen3 7-DoF arm. The output plans generated by INSAT for the experiments (a) and (b) are displayed in Fig. \ref{fig:swinga}, (c) in Fig. \ref{fig:swingb}, (d) in Fig. \ref{fig:shelfsim}, and (e) in Fig. \ref{fig:flip_level}. To minimize torque and spend the least effort by exploiting contact, the manipulator exhibits swinging behavior to reach the contact locations and postures in simulation scenarios (a), (b) and (c). Fig. \ref{fig:torqsw} visualizes the planner's output torque trajectory for the scenario (c) along with the torque reduction ratio (TRR) \cite{braceadv}. $\text{TRR}=(\norm{\boldsymbol{\tau}_{\text{wo}}} -\norm{\boldsymbol{\tau}_{\text{c}}})/\norm{\boldsymbol{\tau}_{\text{c}}}$ where $\boldsymbol{\tau}_\text{wo}$ is the net joint torque from the baseline and $\boldsymbol{\tau}_\text{c}$ is the net torque from INSAT (Eq. \ref{eq:nettorq}). As our baseline cannot discover and exploit contact, the value of $\boldsymbol{\tau}_\text{wo}$ is higher which explains a high TRR of 0.78.  We also found that INSAT leverages passive dynamics as much as possible. Note from Fig. \ref{fig:torqsw} that actuators 3, 5, 6 and 7 remain shut off with zero torque throughout the \textit{slide out} and \textit{swing} phase and activate only during the \textit{land} phase (Fig \ref{fig:swingb}). This suggests that the planner took advantage of the passive dynamics to effortlessly slide out of the confined shelf just using the actuators 2 and 4.

\subsubsection*{Flipping a Bulky Plank}
In some scenarios, it may be undesirable for the body of the robot to come into direct contact with the environment. For example, in the experiments shown in Fig. \ref{fig:shelfhw}, despite our best efforts to protect the robot using bubble wrap, some chips of paint were taken off of the robot's body during contact with the cabinet. However, in these scenarios, the benefits of our method can still be realized by allowing the \emph{payload} to brace against the environment. 

\begin{figure*}%
\centering
\begin{subfigure}{0.66\columnwidth}
\includegraphics[width=\columnwidth]{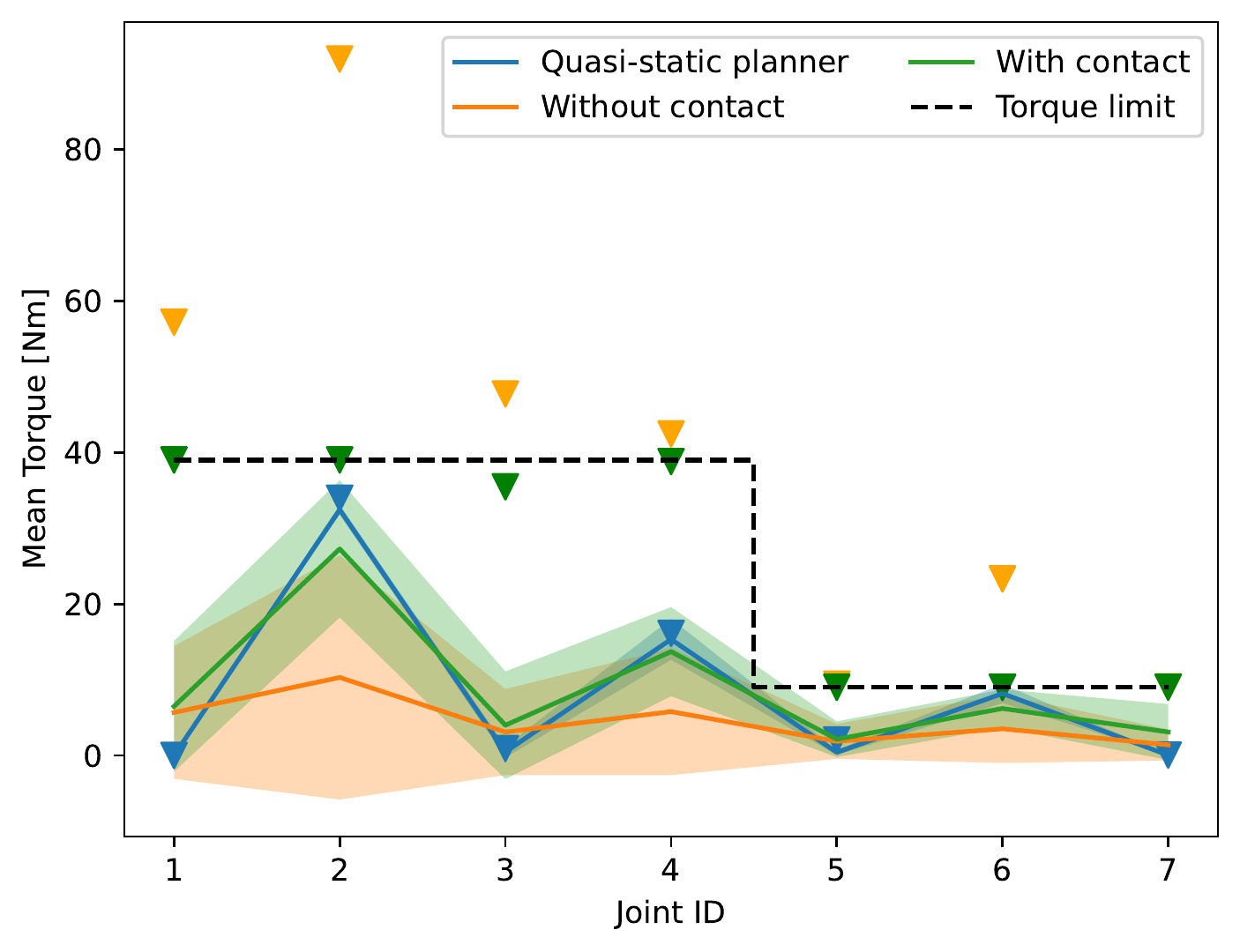}%
\caption{Easy}
\label{fig:sub1}
\end{subfigure}\hfill%
\begin{subfigure}{0.66\columnwidth}
\includegraphics[width=\columnwidth]{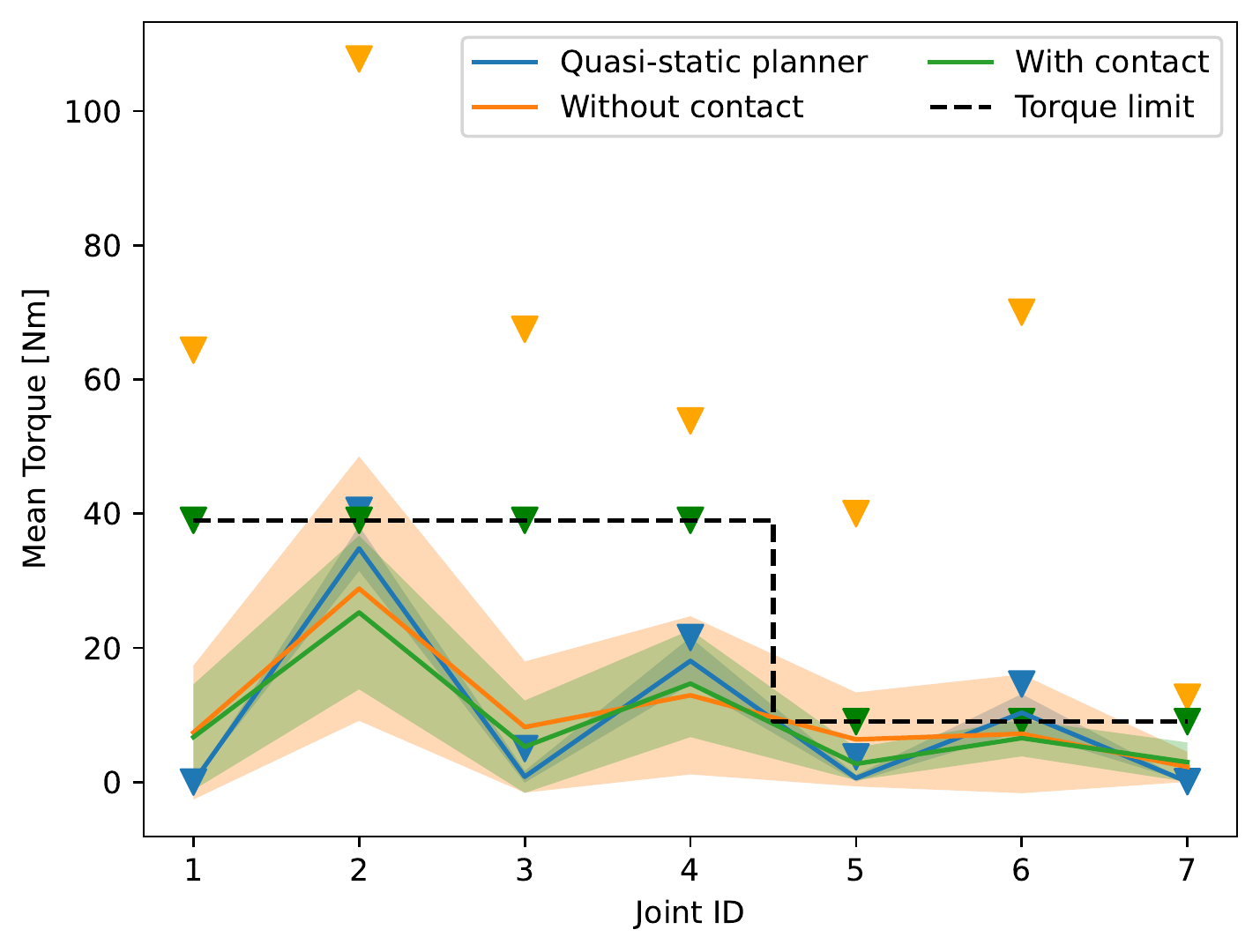}%
\caption{Medium}
\label{fig:sub2}
\end{subfigure}\hfill%
\begin{subfigure}{0.66\columnwidth}
\includegraphics[width=\columnwidth]{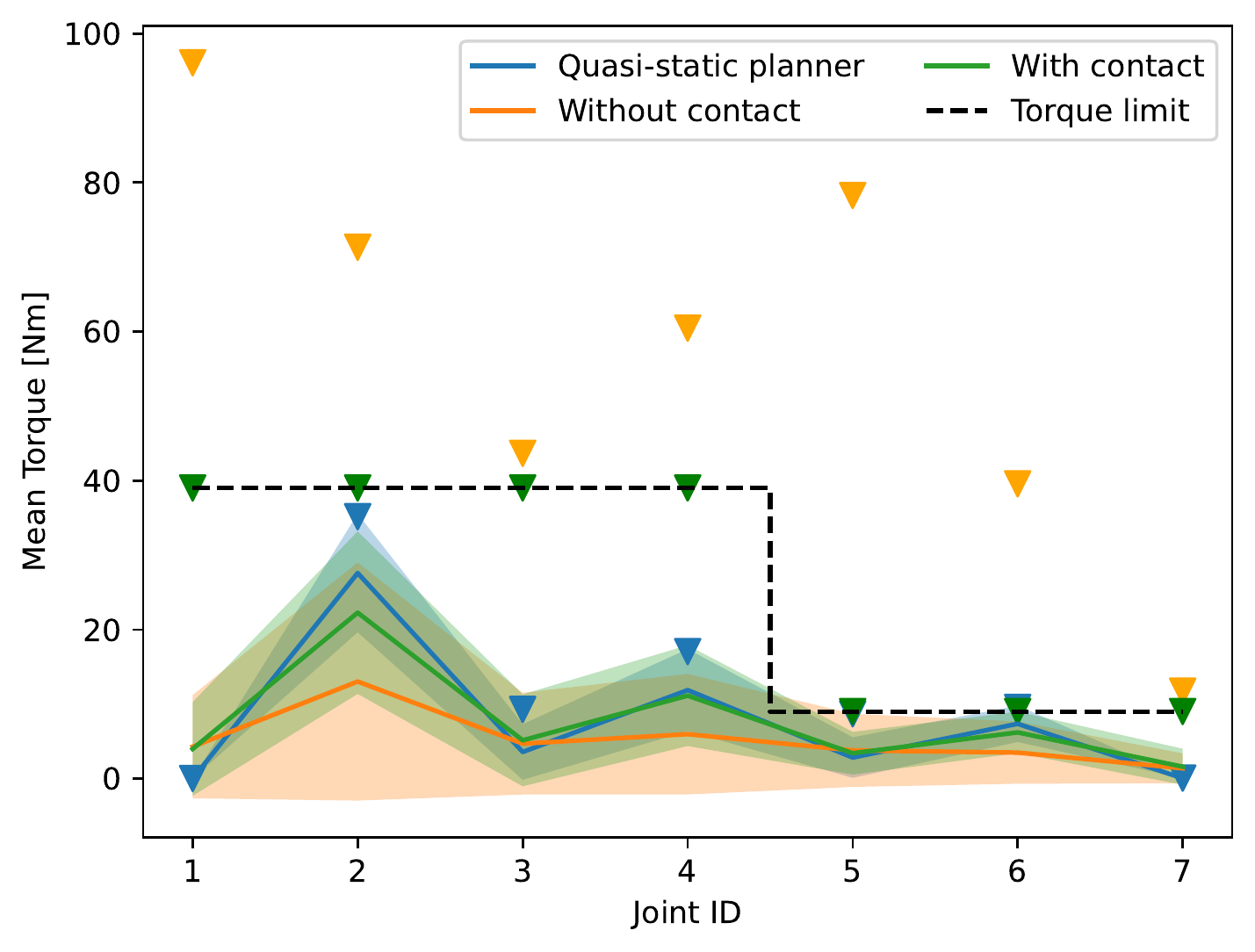}%
\caption{Hard}
\label{fig:sub2}
\end{subfigure}\hfill%
\caption{Comparison of mean absolute torque across joints between our method and the baselines for different difficulty levels of the payload flipping task in simulation. Only the successful runs were used to measure this statistic. Our method is shown in green (with contact) and orange (without contact). The torque values without contact represent the torque required to follow the same trajectory with no environmental support. The colored inverted triangle is the absolute maximum torque of the respective method. Note that only our method does not violate the torque limit across difficulty levels. The lower maximum value for the quasi-static planner in a few instances is attributed to its lower success rate and the quasi-static assumption.}
\label{fig:difftorqplot}
\end{figure*}

\begin{figure*}
\begin{subfigure}{0.5\columnwidth}
\includegraphics[width=\columnwidth]{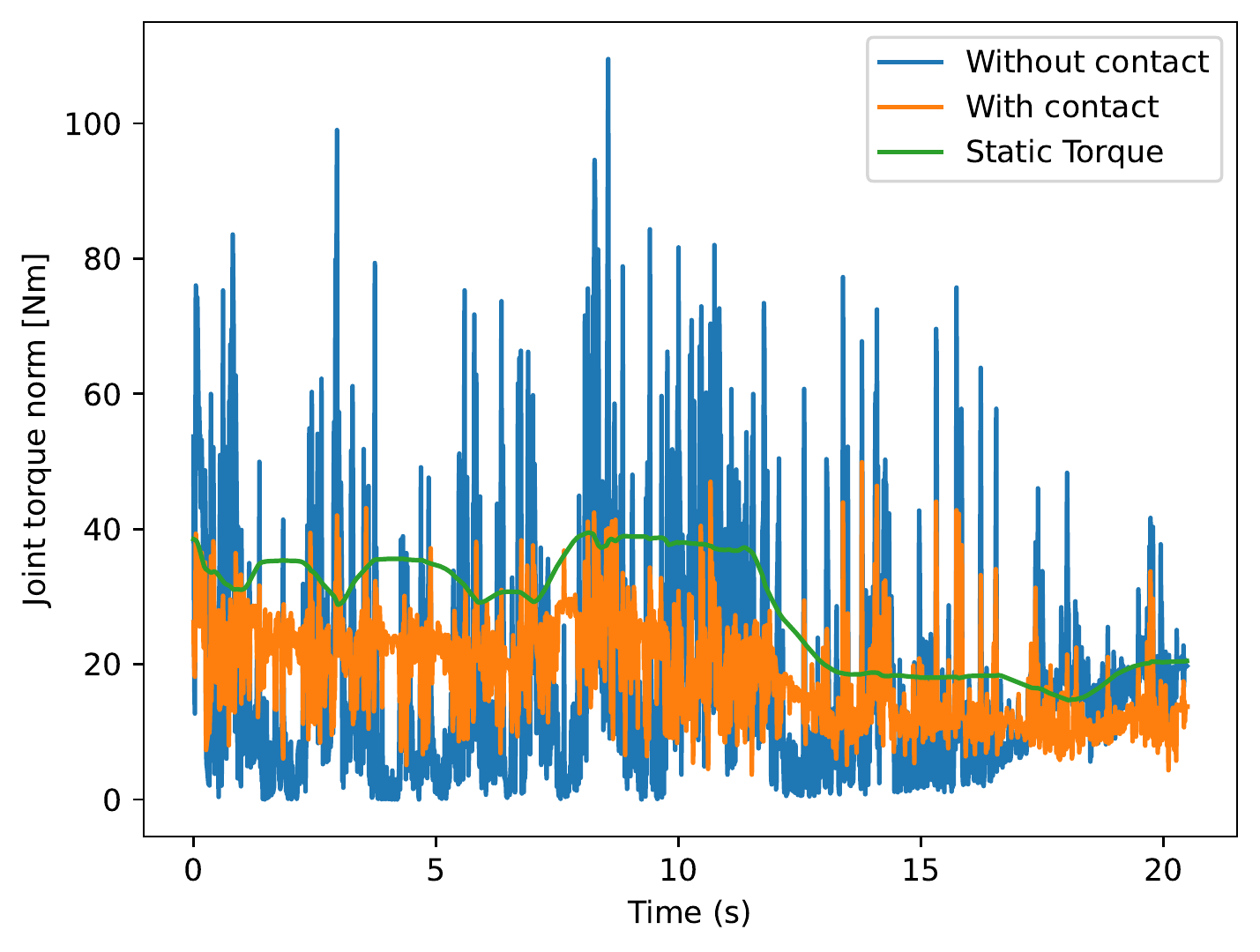}
\end{subfigure}\hfill%
\begin{subfigure}{0.5\columnwidth}
\includegraphics[width=\columnwidth]{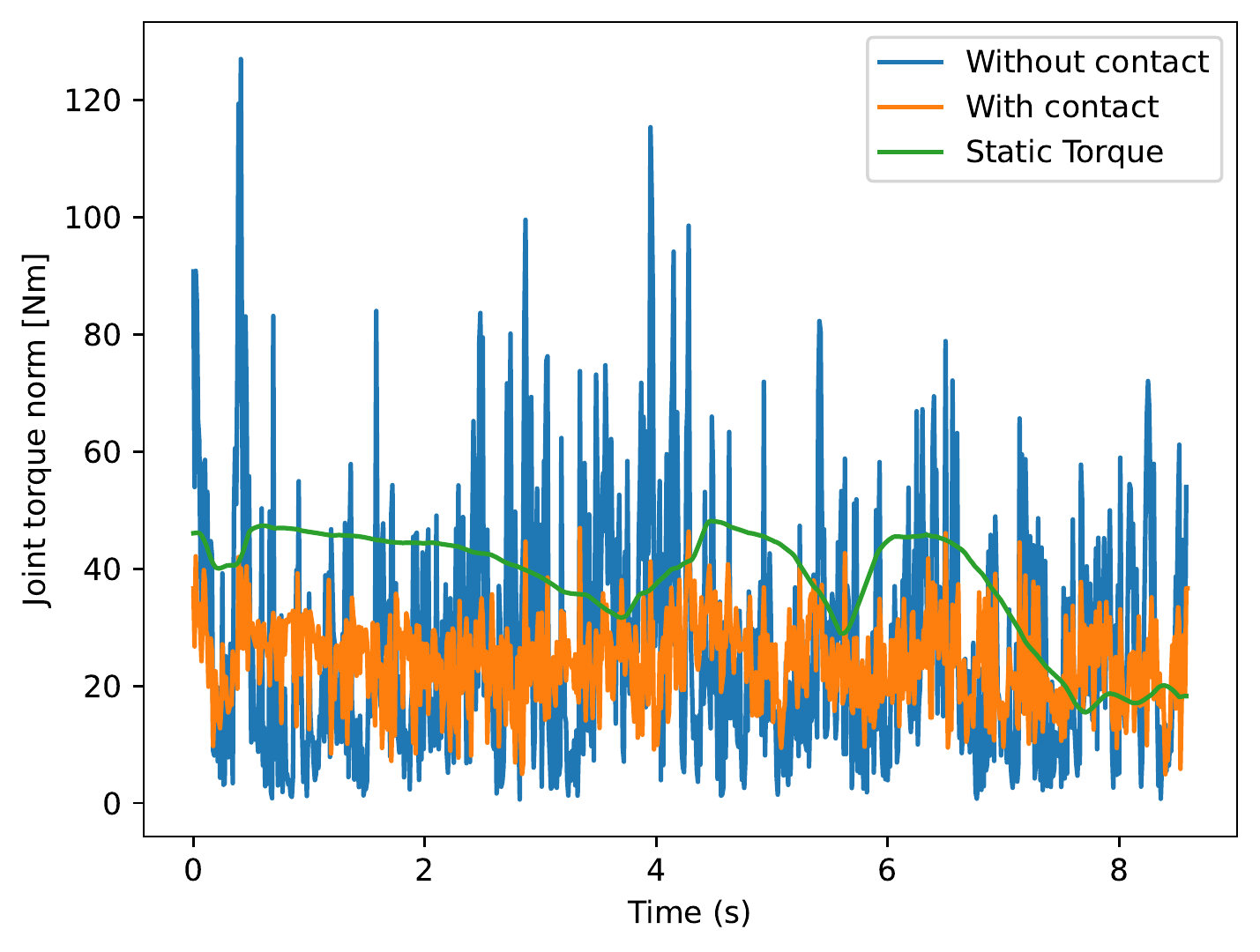}
\end{subfigure}\hfill%
\begin{subfigure}{0.5\columnwidth}
\includegraphics[width=\columnwidth]{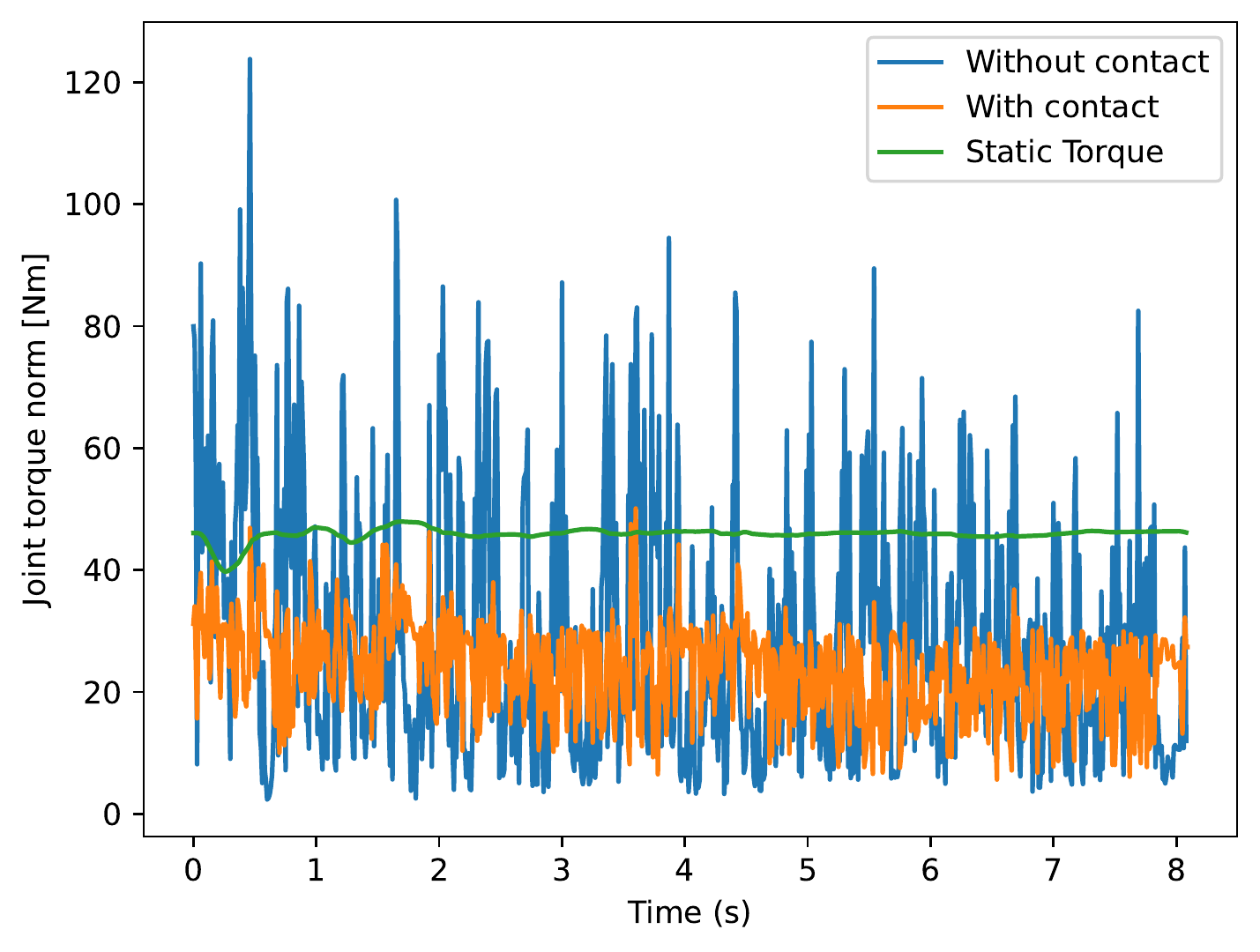}
\end{subfigure}\hfill%
\begin{subfigure}{0.5\columnwidth}
\includegraphics[width=\columnwidth]{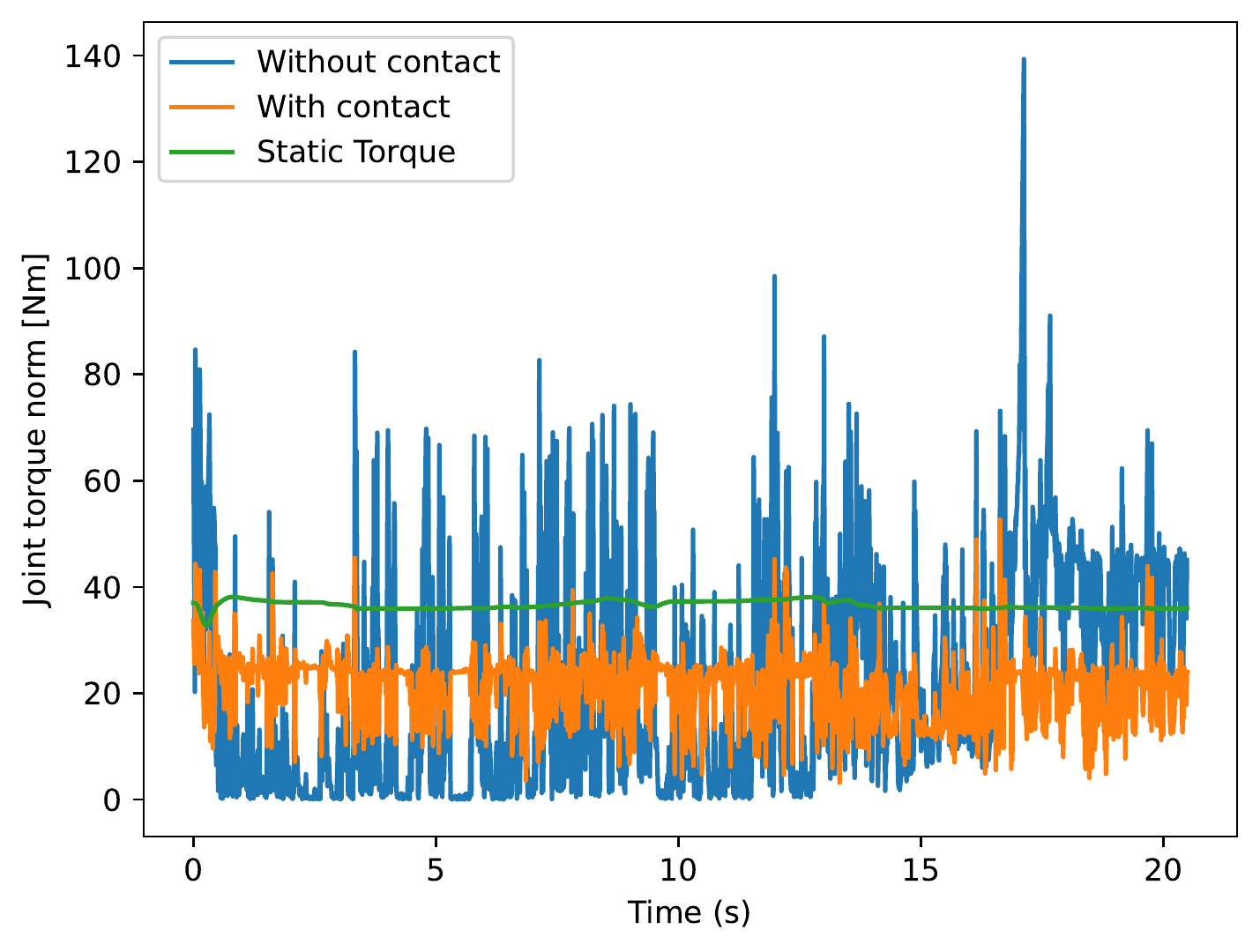}
\end{subfigure}\hfill%

\begin{subfigure}{0.5\columnwidth}
\includegraphics[width=\columnwidth]{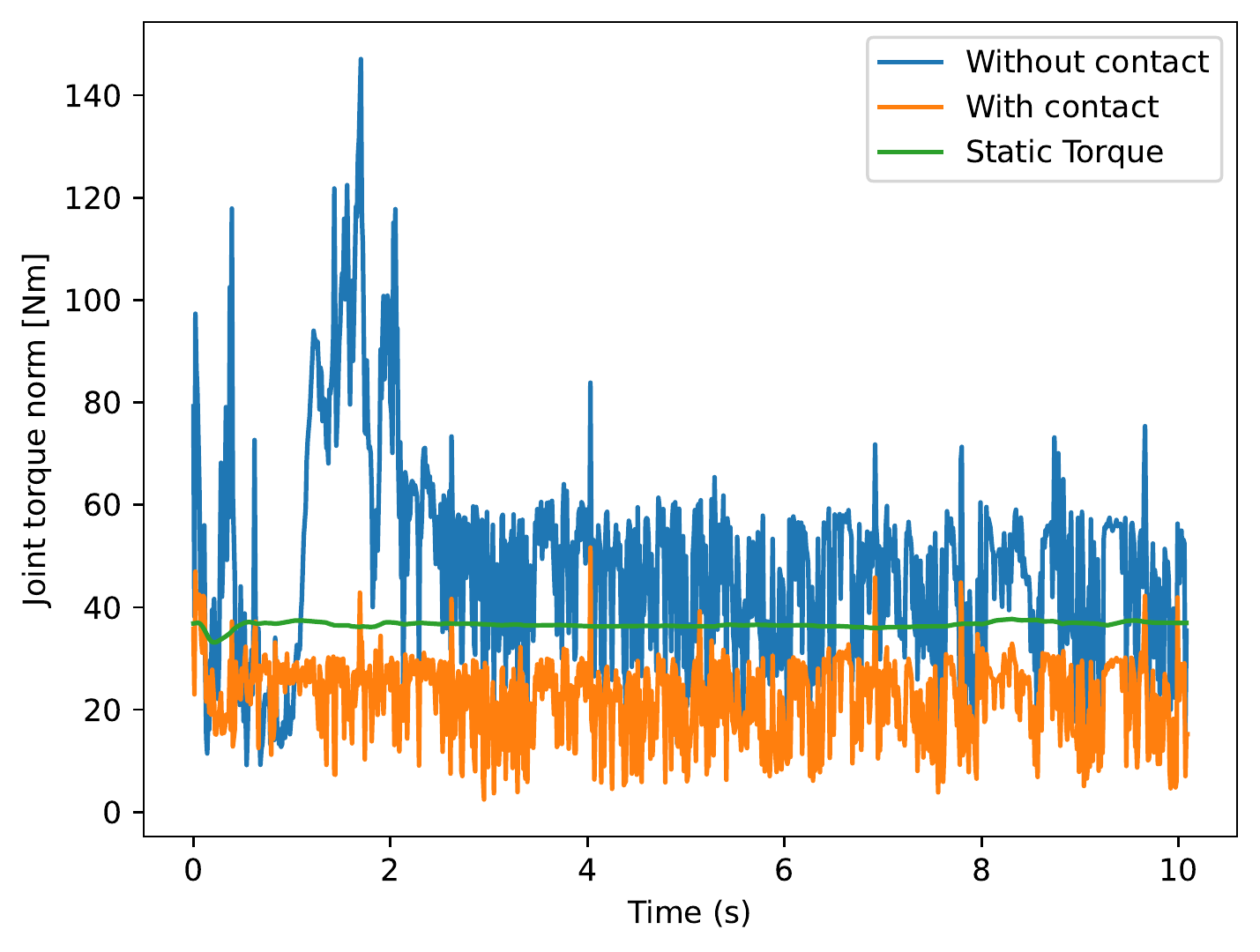}
\end{subfigure}\hfill%
\begin{subfigure}{0.5\columnwidth}
\includegraphics[width=\columnwidth]{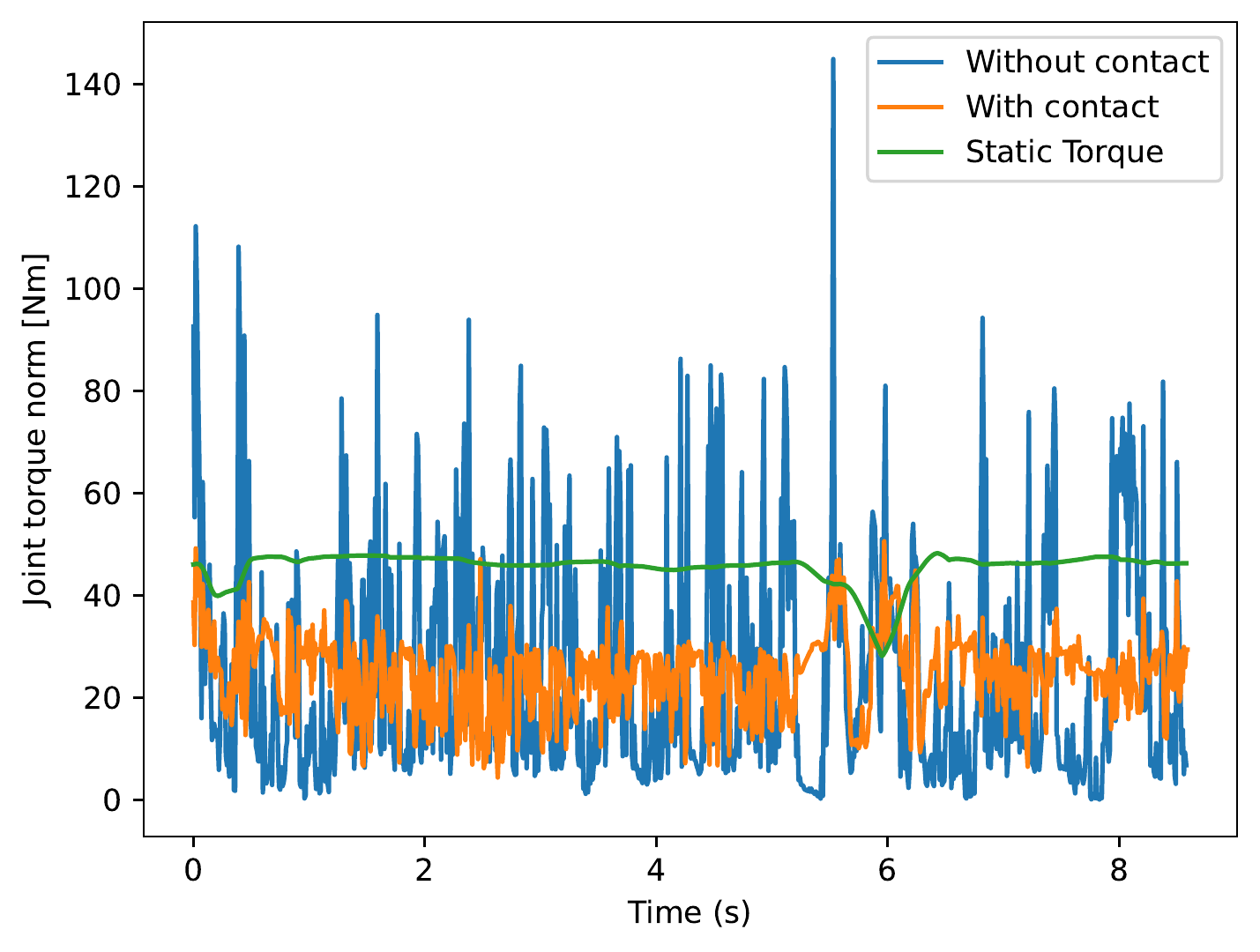}
\end{subfigure}\hfill%
\begin{subfigure}{0.5\columnwidth}
\includegraphics[width=\columnwidth]{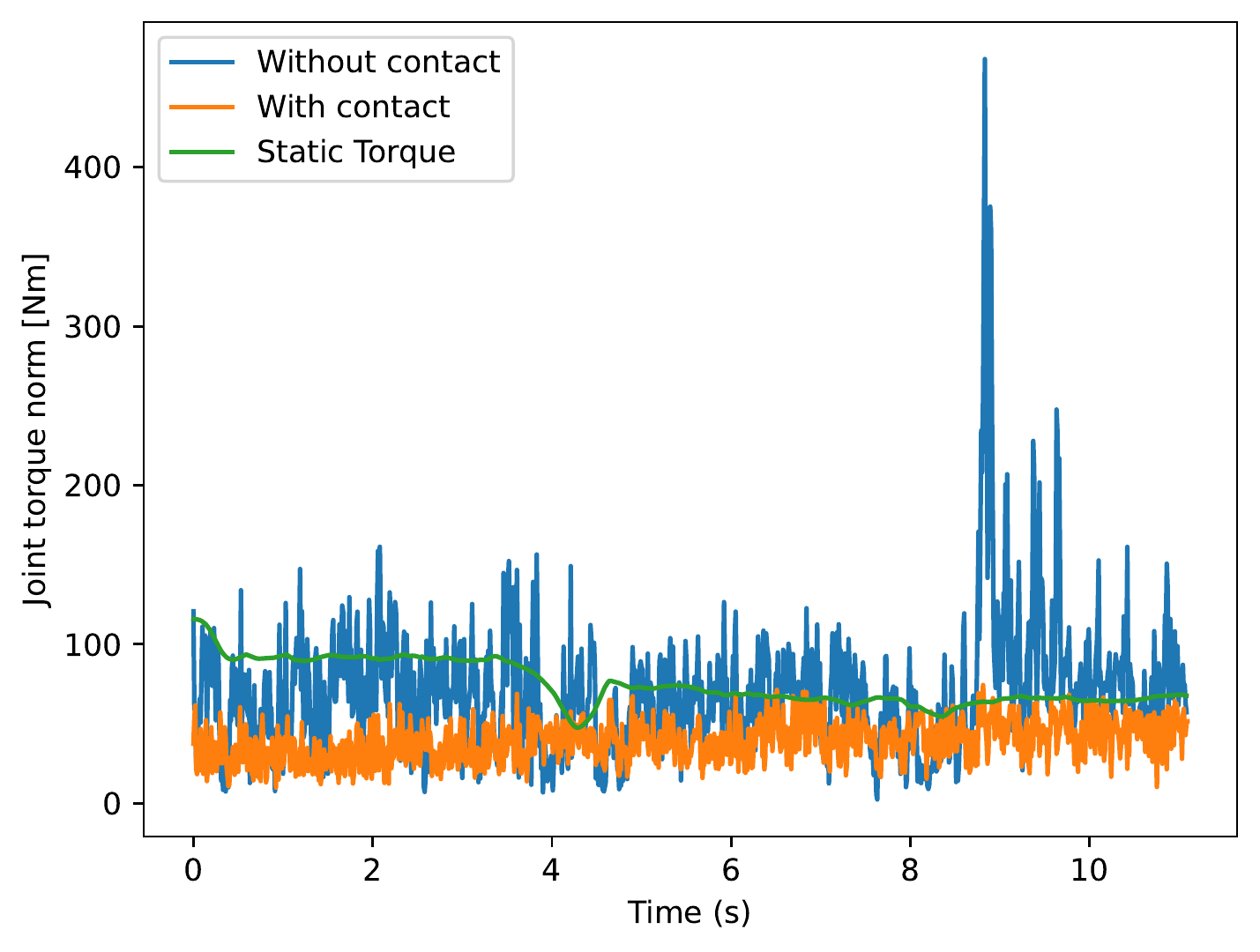}
\end{subfigure}\hfill%
\begin{subfigure}{0.5\columnwidth}
\includegraphics[width=\columnwidth]{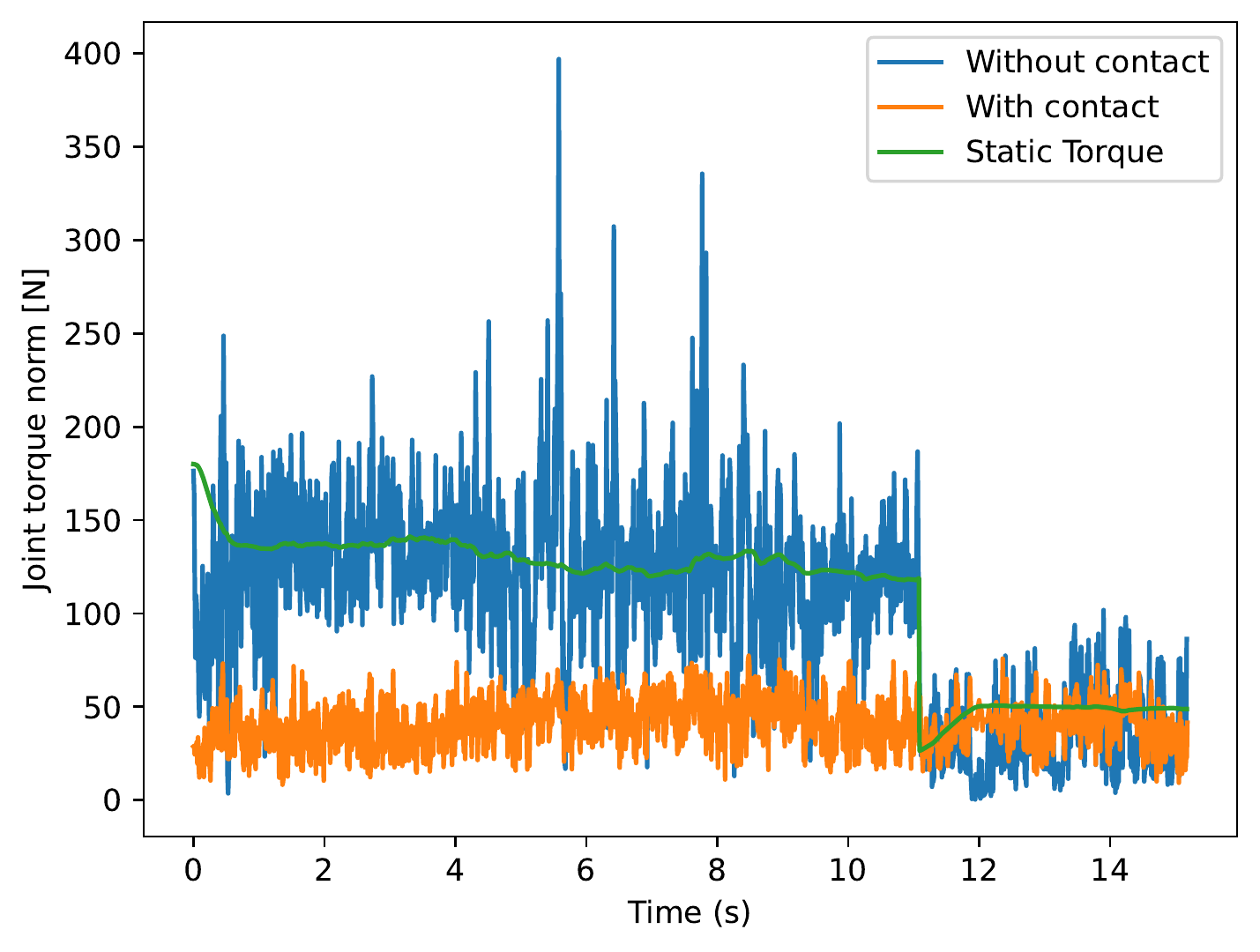}
\end{subfigure}\hfill%
\caption{Time series plots of joint torque norm computed by the planner for different masses of the payload in the medium difficulty payload flipping task. In ROWMAJOR order: the payload is increased from 0.5kg in the top-right to 4kg in the bottom-right with 0.5kg increments. Note that INSAT discovers trajectories with significantly lower net joint torque (orange). The torque values without contact (blue) are the effort required to track the trajectories discovered by INSAT if the environment support was not present. The solution from quasi-static post-processed by trajectory optimization is shown in green.}
\end{figure*}

\par As an example of this application scenario, we placed the Kinova Gen3 robot behind a wall as shown in Fig. \ref{fig:flip_level}-\circled{1}. This wall intersected another wall (henceforth called the partition) at ninety degrees forming a T-shape as shown in Fig. \ref{fig:flip_level}-\circled{2}. The Kinova Gen3 robot was given the task of carrying an overweight plank (Fig. \ref{fig:flip_level}-\circled{3}) from one side of the partition to the other while flipping the plank 180$^\circ$ about its long axis. Figure \ref{fig:flip_level}-\circled{4} shows a semi-transparent version of the goal configuration of the payload. 

\par Despite the overweight nature of the payload and complexity of the manipulation task, our planning method was able to find a trajectory that moved the plank from the start to goal without violating the joint torque limits by utilizing contact between the payload and the wall/partition. Additionally, we were able to vary the height of the partition to increase the difficulty of the task. Three representative partition heights are shown in Fig. \ref{fig:flip_level_2}. Our planner converged for all three cases in simulation and Fig. \ref{fig:flip_level} shows a filmstrip of the simulation for the medium partition height (i.e., the center frame of Fig. \ref{fig:flip_level_2}). 

A total of 10 different masses of the plank is considered for each difficulty level. Starting from the plank weighing 0.5kg, we evaluated INSAT on this task up to 5kg with 0.5kg increments. The resulting distribution of the absolute value of torques (mean and standard deviation) of the trajectories from INSAT across joints is presented in Fig. \ref{fig:difftorqplot}. Note that when successful, INSAT does not violate the torque limit across different masses of the payload (green) although a higher torque is required in every instance to perform the same task without leveraging the environment contact (orange). The time series plot of joint torques for all the instances that were successfully solved in medium difficulty (0.5kg to 4kg) is shown in Fig. \ref{fig:difftorqplot}. Once again it is clear that the torque required by INSAT is the lowest as it actively reasons about contact to stay within the torque limit while completing the task.

\begin{figure}
\centering
\begin{subfigure}{0.8\columnwidth}
\includegraphics[width=\textwidth]{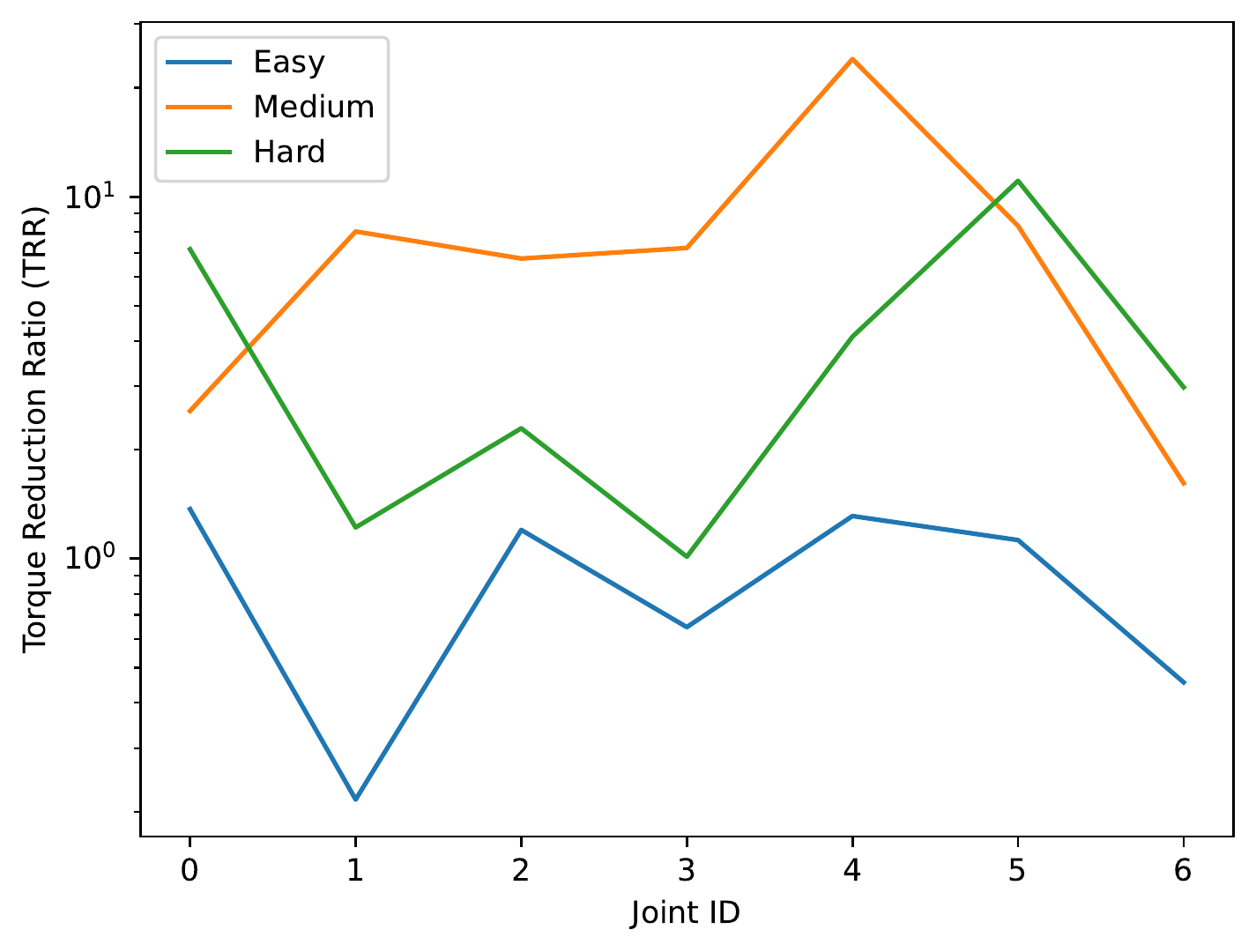}
\caption{Torque Reduction Ratio (TRR) over joints of the Kinova Gen3 arm for different difficulty levels of flipping a bulky payload.}
\label{fig:fliptrr}
\end{subfigure}
\begin{subfigure}{0.8\columnwidth}
\includegraphics[width=\textwidth]{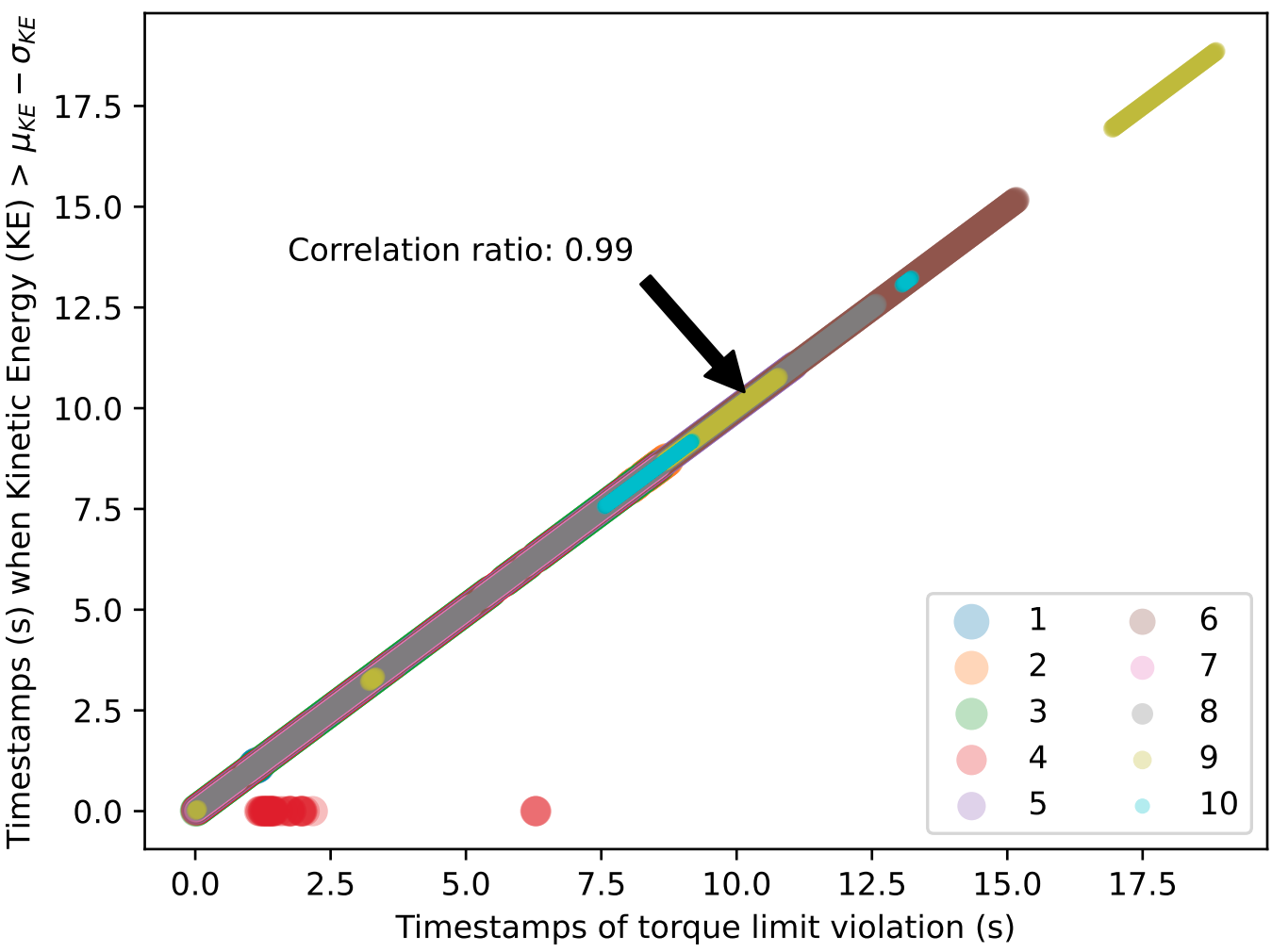}
\caption{The strong correlation of 0.99 indicates a significant relationship between the instances when the torque limit was exceeded in the baseline and the moments when the kinetic energy of the system fell more than one standard deviation below the mean kinetic energy. This suggests that INSAT efficiently managed torque violations in the baseline by directing energy into the system and strategically utilizing it to achieve the desired contact locations.} 
\label{fig:flipcorr}
\end{subfigure} 
\caption{TRR across difficulty levels and analysis of torque violation instances for flipping a bulky plank task.}
\end{figure}

A comprehensive analysis was conducted across 10 distinct masses of the plank for each difficulty level. The evaluation of the INSAT encompassed planks ranging from 0.5kg to 5kg, with increments of 0.5kg. The resultant joint-wise distribution of absolute torque values (including mean and standard deviation) of the trajectories generated by INSAT is illustrated in Figure \ref{fig:difftorqplot}. It is noteworthy that in successful instances, INSAT adheres to the torque limit across different masses of the payload (green), although a higher torque is required in every instance to perform the task without leveraging environmental contact (orange). Figure \ref{fig:difftorqplot} also features a time series plot showcasing joint torques for all instances successfully solved under medium difficulty conditions (0.5kg to 4kg). The graphical representation underscores that INSAT consistently requires the least amount of torque. This efficiency is attributed to its active consideration of contact dynamics, allowing it to strategically operate within the defined torque limits while accomplishing the assigned task. 


The joint-wise Torque Ratio Reduction (TRR) is depicted in Fig. \ref{fig:fliptrr} for all difficulty settings, where a higher TRR indicates that the baseline required more torque to track the same trajectory computed by INSAT. Medium-difficulty tasks exhibit a higher TRR compared to hard tasks, given the lower success rate for hard tasks which are excluded from TRR calculations. Furthermore, Fig. \ref{fig:flipcorr} illustrates the correlation between timestamps when the torque limit was breached in the baseline and timestamps when the kinetic energy of the system is more than one standard deviation below the mean kinetic energy. The substantial correlation of 0.99 suggests that INSAT effectively addressed torque violations in the baseline by channeling energy into the system and strategically utilizing it to reach the appropriate contact locations.

\subsection{Hardware Experiments}
To experimentally validate our method, we planned a trajectory for a Kinova Gen 3 robot to lift a 2.5 Kg payload between adjacent cabinets as shown in Fig. \ref{fig:shelfhw}. The robot was commanded in joint velocity mode using ROS Melodic on Ubuntu 18.04. As a comparison point, the same trajectory was executed in free space (i.e., without the cabinets). During both experiments, the joint torque was recorded using the Kinova Gen3's integrated torque sensors. Table \ref{tab:torque} shows the RMS of the sensed joint torques in both experiments. From these values, it is clear that our planner was able to utilize bracing contacts to meaningfully reduce the torque in most joints when compared to free-space motion.   

\begin{table}[ht]
\renewcommand{\arraystretch}{1.5}
\centering
\resizebox{\columnwidth}{!}{%
\begin{tabular}{p{1.2cm}|ccccccc|c}\toprule
\textbf{Joint ID} &  \textbf{1} & \textbf{2} & \textbf{3} & \textbf{4} & \textbf{5} & \textbf{6} & \textbf{7} & \textbf{Total} \\ \hline\hline
\textbf{Free-space} & 1.93 & 33.28 & 10.14 & 14.06 & 0.97 & 6.29 & 0.20 & 66.86\\\hline
\textbf{Bracing} & 4.65 & 23.96 & 6.67 & 11.80 & 1.28 & 5.67 & 0.59 & 54.62 \\\hline\hline
\textbf{Difference} & -2.72 & 9.32 & 3.46 & 2.26 & -0.31 & 0.63 & -0.40 & 12.24 \\
\bottomrule
\end{tabular}
}
\caption{Experimental RMS torques [Nm] during (i) the braced trajectory shown in Fig. \ref{fig:shelfhw} and (ii) the same trajectory running in free-space (i.e. without the cabinets) producing net savings of 12.24Nm.}\label{tab:torque}
\end{table}

\subsubsection*{Flipping a Bulky Plank}
To further demonstrate the utility of our method for real-world manipulation tasks, we built a physical version of the plank-flipping setup using wood as shown in Fig. \ref{fig:realflip_intro}. The plan was generated offline and executed on the Kinova Gen3 robot in joint velocity mode using ROS melodic running on Ubuntu 18.04. During the trajectory, the joint torques were recorded using the Kinova Gen3's built-in joint torque sensors. The RMS torque for each joint is shown in Table \ref{tab:torque_flip}. Additionally, we ran the same trajectory in free-space (i.e. without the wall and partition) and recorded the joint torques. As shown in Table \ref{tab:torque_flip}, bracing was able to produce a net savings of 34.85 Nm when comparing the RMS torques of the braced experiment to the un-braced experiment.       


\begin{table}[ht]
\renewcommand{\arraystretch}{1.5}
\centering
\resizebox{\columnwidth}{!}{%
\begin{tabular}{p{1.2cm}|ccccccc|c}\toprule
\textbf{Joint ID} &  \textbf{1} & \textbf{2} & \textbf{3} & \textbf{4} & \textbf{5} & \textbf{6} & \textbf{7} & \textbf{Total} \\ \hline\hline
\textbf{Free-space} & 6.12 & 37.73 & 11.05 & 16.8 & 3.11 & 10.25 & 0.48 & 85.53\\\hline
\textbf{Bracing} & 8.61 & 15.5 & 3.39 & 10.71 & 4.94 & 5.11 & 2.41 & 50.68 \\\hline\hline
\textbf{Difference} & -2.48 & 22.22 & 7.65 & 6.09 & -1.83 & 5.13 & -1.93 & 34.85 \\
\bottomrule
\end{tabular}
}
\caption{Experimental RMS torques [Nm] during (i) the braced trajectory shown in Fig. \ref{fig:realflip_intro} and (ii) the same trajectory running in free-space (i.e. without wall and partition) producing net savings of 34.85Nm.}\label{tab:torque_flip}
\end{table}

\section{Conclusion \& Future Directions}

In our previous work, we presented an approach that interleaved low-D graph search and full-D trajectory optimization for manipulation through bracing. We demonstrated that planning with torque and obstacle constraints can be achieved in a manner that identifies bracing locations in the environment, making an otherwise inaccessible configuration reachable due to the torque reduction achieved by bracing. Experiments showed that utilizing bracing contacts can reduce the required actuator torque for a given trajectory. The primary limitation was the planning time, which took approximately 15-20 minutes for complex scenarios (Fig. \ref{fig:shelfhw} and \ref{fig:realflip_intro}). By introducing lazy edge evaluation in INSAT, along with reduced optimization rejection, our planning time has been reduced to an average of 3 minutes, down from over 15 minutes in \cite{insat_ptc}. Our recent work on the multi-threaded version of INSAT (currently under submission), using parallelized graph search algorithms, indicates a 15x further reduction in planning time.

We reiterate that planning with torque and obstacle constraints can be accomplished in a way that identifies bracing locations in the environment, enabling access to configurations that would otherwise be inaccessible due to the torque reduction achieved by bracing. Experiments confirm that the use of bracing contacts can reduce the required actuator torque for a given trajectory. While this work utilizes a full dynamics planner, it's worth noting that some confined spaces may be too restricted to execute any dynamic behavior. Therefore, in future work, we will explore the minimum dynamic model fidelity needed for torque-limited manipulation through contact in confined spaces.

\label{sec:conc}

\bibliographystyle{IEEEtran}
\bibliography{IEEEabrv,mybibfile}

\end{document}